\pdfoutput=1

\documentclass[11pt]{article}

\usepackage[final]{acl}

\usepackage{times}
\usepackage{latexsym}

\usepackage[T1]{fontenc}

\usepackage[utf8]{inputenc}

\usepackage{microtype}

\usepackage{inconsolata}

\usepackage{graphicx}

\usepackage{pdflscape}

\usepackage{url}            
\usepackage{booktabs}       
\usepackage{amsfonts}       
\usepackage{nicefrac}       
\usepackage{xcolor}         
\usepackage{comment}
\usepackage{multirow}
\usepackage{xspace}
\usepackage{subcaption}
\usepackage{pifont}
\usepackage{ragged2e}
\usepackage{enumitem}
\usepackage{longtable}
\usepackage{color, colortbl}

\usepackage{tcolorbox}
\usepackage{enumitem}

\newtcolorbox{promptbox}[1]{
  colback=gray!5,
  colframe=black,
  title=\textbf{#1},
  fonttitle=\bfseries
}

\newtcolorbox{promptbox*}{
  colback=gray!5,
  colframe=black!40,
  boxrule=0.5pt,
  arc=4pt,
  left=6pt,
  right=6pt,
  top=6pt,
  bottom=6pt
}

\usepackage[table]{xcolor}
\usepackage{tikz}
\usetikzlibrary{shapes.geometric, arrows.meta, positioning}

\usepackage{float} 

\usepackage{textcomp}

\definecolor{Gray}{gray}{0.85}
\definecolor{LightCyan}{rgb}{0.88,1,1}
\newcolumntype{a}{>{\columncolor{LightCyan}}r}

\title{Toward Cultural Alignment: Human-Centered Evaluation of Multimodal AI Stories Across Five African Communities}

\author{First Author \\
  Affiliation / Address line 1 \\
  Affiliation / Address line 2 \\
  Affiliation / Address line 3 \\
  \texttt{email@domain} \\\And
  Second Author \\
  Affiliation / Address line 1 \\
  Affiliation / Address line 2 \\
  Affiliation / Address line 3 \\
  \texttt{email@domain} \\}

\author{%
 Millicent Ochieng$^{1}$\thanks{\hspace{0.1cm} Equal Contribution.}, Felermino D. M. A. Ali$^{1}$\footnotemark[1], Elizabeth A. Ankrah$^{1}$, \\
 \textbf{Najeeb G. Abdulhamid$^{1}$, Boyd Migisha$^{2}$, Stephanie Nyairo$^{1}$, Mercy Muchai$^{1}$,}\\
 \textbf{Samuel Maina$^{1}$, Aditya Vashistha$^{3}$, Anja Thieme$^{4}$, Jacki O’Neill$^{1}$}  \\ \\
\footnotesize
$^1$Microsoft Research Africa, 
$^2$Swansea University, 
$^3$Cornell University,
$^4$Microsoft Research Cambridge\\
}

\begin{document}
\maketitle

\begin{abstract} 
In this paper, we examine how well AI-generated multimodal stories align with the lived practices, relationships, language, values, and visual expectations of the communities they represent. We conduct a community-grounded mixed methods evaluation with 19 culture representatives across five African communities, combining quantitative annotations with qualitative focus group discussions. We find that cultural alignment depends not simply on recognizable cultural markers, but on how those markers fit social, linguistic, procedural, and visual context. From these evaluations, we develop a taxonomy of cultural alignment comprising five broader cultural marker categories and eight recurring mechanisms of misalignment. We additionally evaluate five multimodal LLM judges to examine whether automated evaluation can approximate community-grounded judgments at scale. Judge reliability and score calibration vary substantially across communities, with no single judge performing consistently across all five settings. These findings motivate community-calibrated evaluation pipelines in which automated judges are validated against community judgments to determine where they can be trusted and where human review remains necessary. 
\end{abstract}

\section{Introduction}
\label{sec:introduction}

\begin{figure}[h!]
    \centering
    \includegraphics[width=1\columnwidth]{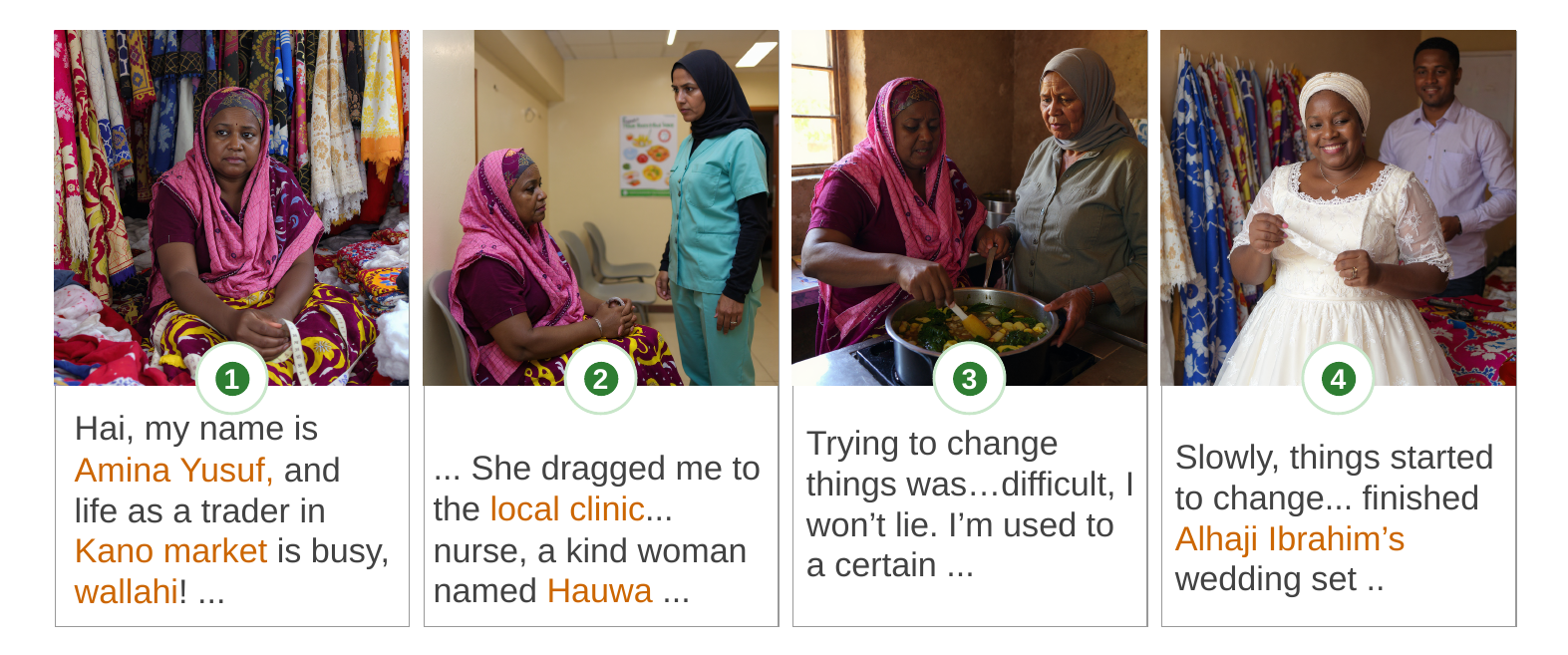}
    \caption{Example story.}
    \label{fig:example_failure}
\end{figure}

Generative AI systems are increasingly used to produce text and visual content for diverse users at scale, including stories~\cite{tian-etal-2024-large-language}. Such content can appear fluent and locally plausible while failing to reflect the lived practices, relationships, language, values, and visual expectations of the communities being represented~\cite{agarwal_fluent_2025,wang-etal-2024-countries,kazemi2024culturalfidelitylargelanguagemodels}. We refer to community-recognized fit between generated content and lived experience as \emph{cultural alignment}. 

Evaluating cultural alignment is difficult because culture is complex, situated, and context-dependent \cite{adilazuarda-etal-2024-towards}. A form of address, item of clothing, or food practice may appear plausible to outsiders while feeling inappropriate, foreign, generic or incomplete to community members. Community-grounded evaluation can surface these situated judgments but is time-intensive to scale across models, communities, and generation settings. LLM-as-judge methods offer a scalable alternative~\cite{li-etal-2025-generation}, but their reliability for culturally situated multimodal evaluation remains uncertain.

We examine cultural alignment through a community-grounded mixed methods evaluation of AI-generated multimodal stories across five African communities: Hausa, Kikuyu, Luo, AmaXhosa, and Xichangana. We combine quantitative annotations and story-level scores from 19 culture representatives with qualitative focus group discussions to examine both what cultural elements shape alignment judgments and why they are experienced as aligned or misaligned.\footnote{We use \emph{culture representatives} to refer to community members who self-identify with the represented community and report relevant lived and linguistic knowledge, enabling them to assess whether generated content feels authentic, inappropriate, foreign, or incomplete.} Narratives are widely used to support behavior-change communication~\cite{hinyard2007narrative,ngendo2026voices}, making stories a useful setting for studying culturally situated generation. Each story consists of four text-image frames, as illustrated in Figure~\ref{fig:example_failure}, and centers on everyday diabetes lifestyle management. We use this domain as a culturally consequential testbed and do not assume that the resulting patterns generalize unchanged to other domains. Stories were generated in English for Hausa, Kikuyu, Luo, and AmaXhosa and in Portuguese for Xichangana, reflecting the role of these languages in formal written communication of the represented communities. 

Because existing annotation tools did not support frame-by-frame evaluation of such multimodal stories across both text and images, we built a custom annotation platform that allowed culture representatives to evaluate each story frame, identify influential text spans and image regions, categorize cultural markers, and provide an overall story-level score. From the annotations and focus group discussions, we developed a taxonomy of cultural alignment that captures the textual and visual markers representatives attend to when judging alignment and the recurring mechanisms through which stories become culturally misaligned. We additionally evaluated five multimodal LLM judges using the same story-level rubric to test whether automated evaluation can approximate community-grounded judgments at scale.

Our analysis yields three main findings. First, our taxonomy shows that cultural alignment is not reducible to the presence of recognizable cultural markers, but depends on how those markers fit context across five broader categories: \textit{Referential} markers such as names, foods, and places; \textit{Procedural} markers such as food preparation and exercise routines; \textit{Contextual} markers concerning when and where cultural elements appear; \textit{Socio-geographic} markers such as clinics, markets, homes, and infrastructure; and \textit{Linguistic Register} markers such as dialect, code-switching, and forms of address. Second, generated stories become culturally misaligned through recurring mechanisms, including substitution, norm violation, omission, forced insertion, register conflation, cross-modal inconsistency, stereotyping, and hallucination. Third, LLM judges do not consistently approximate evaluations from culture representatives; reliability varies by community, and no single judge model performs consistently across all five communities. For example, Pearson correlations between LLM judge scores and scores from culture representatives are strong for Luo ($r=0.82$--$0.89$) and Hausa ($r=0.67$--$0.80$), but no AmaXhosa judge remains significant after correction, and for Xichangana, judges assign alignment scores 35--45 points higher than culture representatives. These findings motivate community-calibrated evaluation in which community judgments ground cultural alignment assessment and automated judges are validated to determine where they can be trusted and where human review remains necessary. Data and code: \url{https://github.com/microsoft/Multimodal-Cultural-Alignment-Africa}.

\section{Related Work}
\label{sec:related_work}

\paragraph{Cultural Alignment and Operationalizing Culture in LLMs.}
A growing body of work probes LLMs for cultural knowledge using proxies such as Hofstede's dimensions~\cite{arora-etal-2023-probing,cao-etal-2023-assessing}, moral judgment datasets~\cite{NEURIPS2023_a2cf225b,jinnai-2024-cross}, and social etiquette norms~\cite{rao-etal-2025-normad}. However, culture is difficult to operationalize because it cannot be reduced to static demographic labels, national categories, or isolated value dimensions~\cite{adilazuarda-etal-2024-towards}. Prior evaluations also show that LLMs often align more strongly with Western cultural norms, and that English prompts can flatten cross-cultural variation~\cite{cao-etal-2023-assessing,agarwal_ai_2025,rao-etal-2025-normad}. While these studies are important for measuring cultural knowledge and bias, knowing about a culture is different from generating content that communities recognize as culturally aligned. Our work builds on this distinction by evaluating whether generated multimodal stories reflect lived practices, relationships, language, values, and visual expectations as judged by culture representatives.

\paragraph{Cultural Misalignment in Generated Text and Images.}
Research on cultural misalignment in generated content has examined failures in dialogue, narrative, and visual generation. In dialogue, NormDial~\cite{li-etal-2023-normdial} and ReNoVi~\cite{zhan-etal-2024-renovi} annotate norm adherence and violation in Chinese and American conversations, while cross-cultural work shows that culture-specific reasoning often fails to generalize~\cite{jinnai-2024-cross}. In narrative generation, prior work has identified Western bias in stories about Arab cultures~\cite{naous2024beer} and introduced taxonomies of cultural misrepresentation for Indian stories~\cite{bhagat2026tales}. In the visual domain, text-to-image (T2I) models have been shown to neglect or misrepresent disadvantaged and underrepresented cultures~\cite{zhang2024partiality,kannen2024beyond,johnson2026,ThiemeEngagingCommunities}, and recent benchmarks characterize \emph{how} these failures arise. CuRe~\cite{rege2025iccv-cure} traces them to the long tail of web-scraped training data, showing that T2I systems hallucinate details for artifacts of the Global South (e.g., an Ethiopian \emph{jebena}) that they render reliably for better-represented ones. CulturalFrames~\cite{nayak-etal-2025-culturalframes} distinguishes \emph{explicit} expectations, stated in the prompt, from \emph{implicit} ones implied by its cultural context, and finds that expectations are missed 44\% of the time across 10 countries, including 68\% of explicit and 49\% of implicit cases. Moving beyond object-centric artifacts, CULTIVate~\cite{malakouti2026culture} evaluates cultural faithfulness through social activities such as dining, greeting, and dance, where meaning emerges from interaction and context rather than isolated objects, and decomposes failure into alignment, hallucination, exaggeration, and diversity, reporting systematically lower faithfulness for Global South than Global North cultures. Notably, the studies find that standard image--text alignment metrics correlate poorly with human judgments of cultural alignment, motivating culturally grounded evaluation. Complementing these automated benchmarks, other work examines community-driven methods for assessing cultural sensitivity~\cite{kiden2025communities}. These studies show that cultural misalignment appears in both language and images, but they evaluate images from short, isolated prompts; less work evaluates text and image jointly within the same culturally situated narrative, or examines the mechanisms through which multimodal stories become misaligned.

\paragraph{Scalable Evaluation and LLM-as-Judge.}
LLM-as-judge methods have been widely adopted as scalable alternatives to human evaluation for assessing generated content~\cite{zheng2023judging,kim2024prometheus,li-etal-2025-generation}. These methods can reduce evaluation cost and increase coverage, but their reliability is sensitive to task framing, model bias, and subjective evaluation criteria~\cite{khan2025randomness}. Cultural alignment evaluation is especially challenging because judgments may depend on lived and linguistic knowledge of the represented community, including whether language use, social norms, visual settings, and cultural markers feel appropriate in context. Our work therefore evaluates multimodal LLM judges against scores from culture representatives rather than treating automated scores as ground truth. In doing so, we examine when automated judges approximate community evaluations, where they are biased, and why community validation remains necessary for scalable cultural alignment evaluation.

\section{Methodology}
\label{sec:methodology}
\begin{figure*}[h!]
    \centering
    \includegraphics[width=1\textwidth]{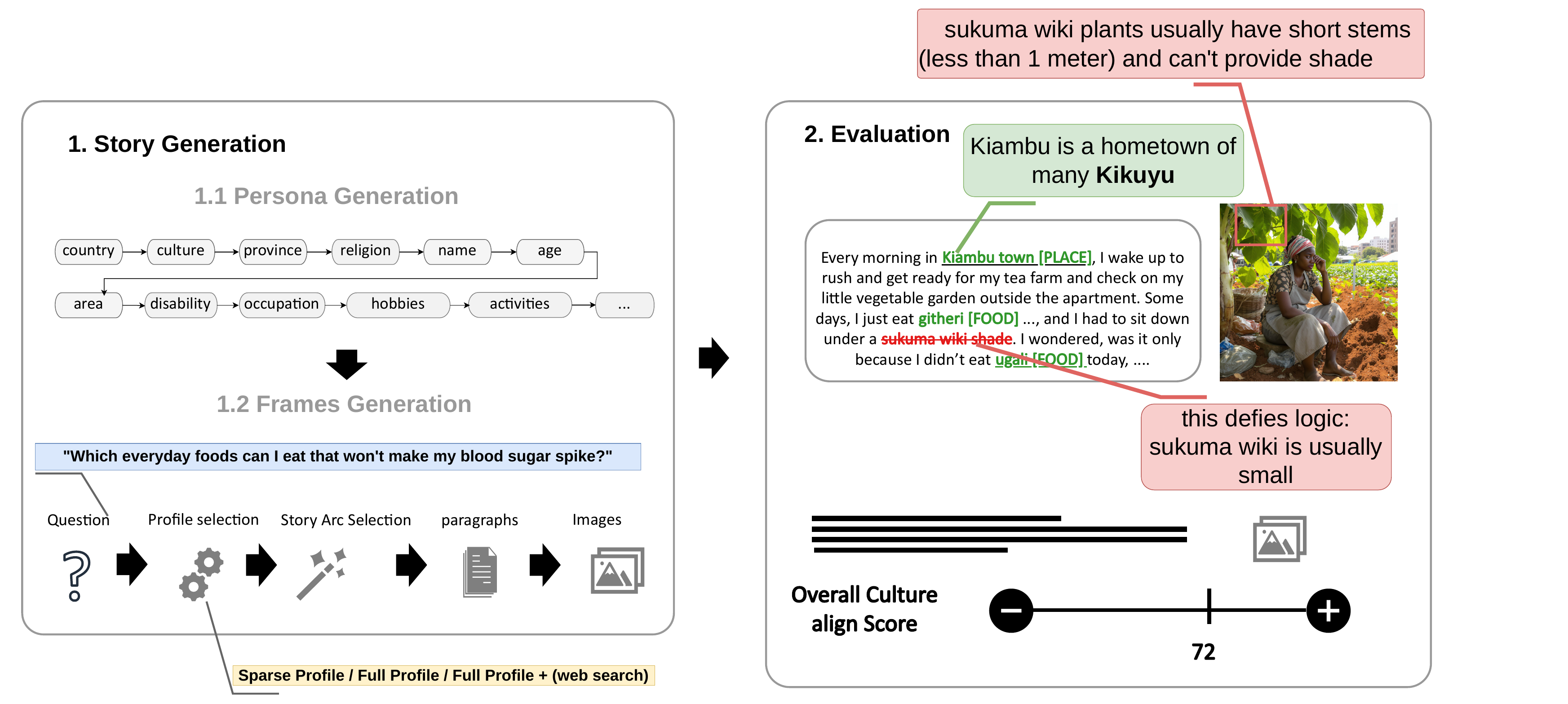}
    \caption{Generation and Evaluation workflow. Green highlights indicate culturally appropriate elements; red highlights indicate violations identified by culture representatives.}
    \label{fig:workflow}
\end{figure*}

We adopt a mixed-methods approach to evaluate cultural alignment in generated multimodal stories, validate automated evaluations against community-grounded judgments, and characterize the cultural evidence and misalignment mechanisms identified by culture representatives. Figure~\ref{fig:workflow} illustrates the end-to-end pipeline from story generation through cultural evaluation, and the following sections describe each stage.

\subsection{Story Generation}
\label{sec:storygeneration}

\paragraph{Persona generation.}
We generated culturally situated personas to provide demographic and cultural context for story generation. Using GPT-4.1, we created personas through a cascading pipeline of 19 sequential LLM calls, with each attribute conditioned on previously generated attributes (e.g., country $\rightarrow$ community $\rightarrow$ province $\rightarrow$ religion $\rightarrow$ name $\rightarrow$ gender $\rightarrow$ age). This dependency-aware process was designed to maintain consistency across demographic, cultural, and regional attributes. The generated personas were manually reviewed for plausibility by authors from the represented communities, and the same reviewed persona pool was reused across all seven generation models. For each community, we generated 40 personas, each with 28 interrelated attributes used as context for story generation; the full set of persona fields is provided in Appendix Table~\ref{tab:profile-fields}. Each persona was paired with an everyday diabetes lifestyle question to guide story generation while keeping the focus on everyday life rather than medical diagnosis or treatment. The questions are listed in Appendix Table~\ref{tab:health-questions}.

\paragraph{Frame generation.}
A story consists of four first-person frames, where each frame contains a short paragraph paired with a corresponding image. Text generation was conditioned on the generated persona, selected diabetes lifestyle question (Appendix Table~\ref{tab:health-questions}), predefined narrative arc (Appendix Table~\ref{tab:story-arcs}), and one of three context settings used to vary the information available to the model during generation (Appendix Table~\ref{tab:context-settings}). The prompt instructed models to write each story in a conversational, first-person style, using English for Hausa, Kikuyu, Luo, and AmaXhosa personas, and Portuguese for Xichangana personas.\footnote{We used English for Hausa, Kikuyu, Luo, and AmaXhosa stories because English is widely used for formal written communication and education in Nigeria, Kenya, and South Africa. We used Portuguese for Xichangana stories because the study context is Mozambique, where Portuguese serves this role.} The system and user prompts used for story generation are provided in Appendix Figures~\ref{prompt:system} and~\ref{prompt:user}. We used seven generation models spanning model families, parameter scales, and access types to produce a diverse set of AI-generated stories across larger and smaller, open-weight and proprietary systems: GPT-4.1, GPT-4.1-mini, Gemma-3 27B, Gemma-3 4B~\cite{gemmateam2025gemma3technicalreport}, Qwen3-4B~\cite{yang2025qwen3technicalreport}, Llama 3.1 8B, and Llama 3.3 70B~\cite{grattafiori2024llama3herdmodels}. For image generation, we used FLUX.1-Kontext-pro~\cite{labs2025flux1kontextflowmatching}. The first image was generated from the first paragraph, and subsequent images were produced through iterative image editing conditioned on the previous frame, using a fixed seed of 12345 and a low editing strength of 0.3 to support character consistency and visual continuity. The final dataset contains 199 multimodal stories across five communities, seven generation models, and three context settings.

\subsection{Story Evaluation}
\subsubsection{Evaluation Procedure}

We designed the evaluation procedure to capture cultural alignment at multiple levels, including the overall story, each text-image frame, and which specific cultural markers within each frame shaped evaluators' judgments. The procedure was informed by prior work on operationalizing culture through demographic and semantic proxies, which we adapted into cultural marker categories for text and image annotation~\cite{adilazuarda-etal-2024-towards}; the full category list is provided in Appendix Table~\ref{tab:marker-categories}. These categories covered markers such as language, cultural group, region, religion, socioeconomic context, age and gender roles, occupation, food and dietary norms, physical activity norms, kinship and social structure, community practices, social etiquette, values and beliefs, and attitudes toward health.

Because existing annotation tools did not support the full workflow required for multimodal cultural alignment evaluation, we designed a custom annotation platform for this study. The platform supports frame-level evaluation of text and images, text-span highlighting, image-region selection, cultural marker categorization, connectedness labels, influence ratings, optional comments, and story-level alignment scoring. These features separate overall judgments from the specific textual and visual evidence behind them, allowing evaluators to indicate not only whether a story felt aligned or misaligned, but which cultural markers shaped that judgment and how strongly. Figure~\ref{fig:platform1} shows the interface used by culture representatives.

At the frame level, evaluators assessed how culturally connected each paragraph and accompanying image felt to the represented community using three response options: \textit{Not connected}, \textit{Connected}, and \textit{Strongly connected}. They then identified the specific text spans or image regions that influenced their judgment, assigned each selected marker a cultural category (see Figure~\ref{fig:platform2}), indicated whether the marker was culturally connected or not connected, rated how strongly it influenced their judgment, and optionally provided a comment. We also provided audio to culture representatives as an accessibility aid for reviewing the story, rather than as an evaluated modality.

After completing all four frames, evaluators completed a story-level feedback checklist covering cultural fit, persona consistency, image reliability, safety, and overall story experience. They then assigned an overall cultural alignment score from 0 to 100 using an anchored rubric (see Figure~\ref{fig:platform3}). The rubric ranged from no recognizable cultural markers or cultural relevance to strong cultural alignment with high relevance, coherence, and authenticity in relation to the represented community (Table~\ref{tab:score-rubric}). We used this graded score because cultural alignment is not binary. A story may contain recognizable cultural markers while still varying in relevance, integration, and multimodal coherence. The same story-level scale was used for LLM judges and culture representatives to support comparison between automated and community-grounded evaluations.

\subsection{LLM Judge Evaluation}
\label{sec:meth-judge}

We evaluated each generated story with LLM judges to test whether automated evaluators can provide a scalable approximation of community judgments of cultural alignment. We used five multimodal LLM judges: Kimi-K2.6, Gemma-4-31B, GPT-5.5, Qwen-3.5-9B, and Qwen-3.5-122B. These models were selected to cover a range of model families, scales, and access types, including proprietary and open-weight systems, while supporting evaluation of text, images, and text-image coherence. No exact generation model was reused as an LLM judge. Each judge evaluated the same story set using the same story-level cultural alignment rubric used by culture representatives. The judge prompts also included the same cultural marker categories used in the community evaluation, but did not include community-specific calibration examples, as our goal was to evaluate whether general rubric-based judges could approximate community judgments without prior community-specific calibration. We conducted three independent evaluation runs per judge to account for variability in model outputs, and averaged the three runs into a single judge score for each story before comparing automated scores with scores from culture representatives. The full judge prompt is provided in Figure~\ref{prompt:llm-as-judge} and Figure~\ref{prompt:judge-user}.

\subsection{Community Evaluation}
\label{sec:meth-evaluation}

\paragraph{Participants and Communities.}
We recruited 19 participants aged 18--44 who served as culture representatives across five African cultural communities: Hausa (Northern Nigeria), AmaXhosa (Eastern Cape, South Africa), Luo (Western Kenya), Kikuyu (Central Kenya), and Xichangana (Maputo, Mozambique), as shown in Appendix Figure~\ref{fig:cultures-map}. Participants were selected based on self-identification with the target community, lived experience in the relevant cultural region, and native or regular use of the relevant community language. They were recruited through community networks, including referrals from local collaborators, professional contacts, and community-based networks connected to the represented cultural groups. All participants provided written informed consent, received an internet allowance before the study to support participation, and were compensated with gift vouchers upon completion. Community context and participant demographics are provided in Appendix~\ref{app:communities} and Appendix Table~\ref{tab:culture-representative-demographics}.

\paragraph{Individual Cultural Evaluation.}
Culture representatives completed a structured onboarding and practice annotation process. We held a one-hour onboarding session to introduce the study goals, define key terminology, explain the cultural marker categories, and demonstrate the annotation platform. Representatives then completed a one-week training phase, during which they annotated 20 practice stories using the same interface and evaluation procedure used in the main study. After this phase, we held a one-hour discussion session to review examples of agreement and disagreement, clarify annotation expectations, and support a shared understanding of the evaluation task while still allowing community-specific interpretations. Culture representatives then individually evaluated 40 stories generated for their own communities using the evaluation procedure described above. In total, the individual evaluation produced 18,805 marker-level annotations across 199 stories, including 9,386 text span annotations (Appendix Table~\ref{tab:text-annotation-stats}) and 9,419 image region annotations (Appendix Table~\ref{tab:image-annotation-stats}).

\paragraph{Focus Group Discussions.}
Following individual annotation, we conducted 15 focus group sessions, with three sessions per community. Each session lasted approximately two hours, for a total of approximately 30 hours of recorded discussion. Sessions were conducted over Microsoft Teams, audio-recorded, and transcribed.

The focus groups were designed to collect qualitative explanations and examine how representatives reasoned through agreement and disagreement in their individual annotations. In the first session, representatives discussed stories where their annotations showed broad agreement, helping establish shared vocabulary for cultural alignment and misalignment within each community. In the second session, representatives examined stories where their annotations diverged, surfacing implicit norms, contested expectations, and differences in how cultural dimensions were weighted. These discussions allowed representatives to revisit details they may have missed during individual annotation, clarify the reasoning behind their judgments, and decide whether to maintain or revise their interpretations. In the third session, representatives reflected across the full set of evaluated stories, identifying the strongest and weakest examples and distinguishing meaningful cultural integration from superficial decoration. The guiding questions are provided in Appendix Table~\ref{tab:focus-group-protocol}. These discussions served as a deliberative evaluation method, producing the situated interpretations from which the taxonomy in Section~\ref{sec:meth-taxonomy} was derived.

\subsection{Taxonomy Derivation}
\label{sec:meth-taxonomy}

\begin{table}[t]
\centering
\scriptsize
\setlength{\tabcolsep}{3pt}
\begin{tabular}{cp{2cm}p{5cm}}
\toprule
&\textbf{Taxonomy} & \textbf{Description and examples} \\
&\textbf{category} &  \\
\midrule
& Referential & Explicit cultural references, such as names, foods, places, objects, and institutions. \\
\multirow[c]{5}{*}[0pt]{\rotatebox{90}{\textbf{Cultural marker categories}}} &Procedural & How everyday practices are carried out, such as food preparation, care-seeking, exercise routines, and household activities. \\
&Contextual & Whether cultural elements fit the situation, timing, purpose, or setting in which they appear. \\
&Socio-geographic & Place-based social and material conditions, such as clinics, markets, homes, infrastructure, mobility, and local environments. \\
&Linguistic register & Community-specific language use, including dialect, code-switching, forms of address, tone, and community voice. \\
\midrule

&Substitution & Markers from another community replace target-community markers. \\
\multirow[c]{6}{*}{\rotatebox{90}{\textbf{ Misalignment mechanisms}}} &Norm violation & Content contradicts community norms, expectations, or respectful conduct. \\
&Omission & Expected cultural markers or situated details are absent, making the story feel generic. \\
&Forced insertion & A plausible cultural marker appears without meaningful integration into the story or context. \\
&Register conflation & The story uses the wrong language variety, dialect, tone, or form of address. \\
&Cross-modal inconsistency & Text and image provide conflicting cultural cues. \\
&Stereotyping & The story or image reduces a community to narrow, repeated, or generic tropes. \\
&Hallucination & The model fabricates or misattributes a cultural practice, object, or meaning. \\
\bottomrule
\end{tabular}
\caption{Summary of the cultural alignment taxonomy derived from focus group discussions and participant annotations (refer to Figure~\ref{fig:cultural-alignment-taxonomy} for examples).}
\label{tab:taxonomy-structures}
\end{table}
  
We developed the cultural alignment taxonomy through a convergent mixed-methods design \cite{creswell2018designing}, conducting thematic analysis \cite{braun2006using,braun2019reflecting} of the FGDs to derive the taxonomy, and then examined how the resulting categories appeared in the participant annotations. Each FGD was conducted with two to three of the authors present, during the session they took independent notes. After the session each transcript  was put into HeyMarvin, a qualitative research tool, for transcription and analysis. Each transcript was anonymised and read in full by at least one author. The authors noted emergent categories from their notes and during their readings, extracted the examples from the transcripts which fell into these categories. Five of the authors then conducted 5 joint analysis sessions where they discussed the categories and examples together, and from these identified emergent themes with verifiable examples. These themes formed the basis of the cultural alignment taxonomy and the recurring misalignment mechanisms summarized in Table~\ref{tab:taxonomy-structures}. The cultural marker categories organize the annotated text and image markers, labeled using the fine-grained categories in Appendix Table~\ref{tab:marker-categories}, into broader categories. The misalignment mechanisms capture recurring failure patterns identified from representatives' reasoning in the focus group discussions.

\section{Results}
\label{sec:results}

\subsection{Community Evaluation Reveals Patterned Cultural Alignment and Misalignment}
\label{sec:results-community-eval}

The uneven judge results show that automated scores alone cannot explain cultural alignment. Community evaluation adds this missing layer by showing which textual and visual markers representatives used to judge cultural fit, and how those markers supported, weakened, or disrupted alignment in context. We organize these results using the taxonomy summarized in Table~\ref{tab:taxonomy-structures}, first examining the five broader cultural marker categories across modalities and then describing the eight recurring mechanisms of misalignment.

\begin{figure*}[h!]
    \centering
    \includegraphics[width=1\textwidth]{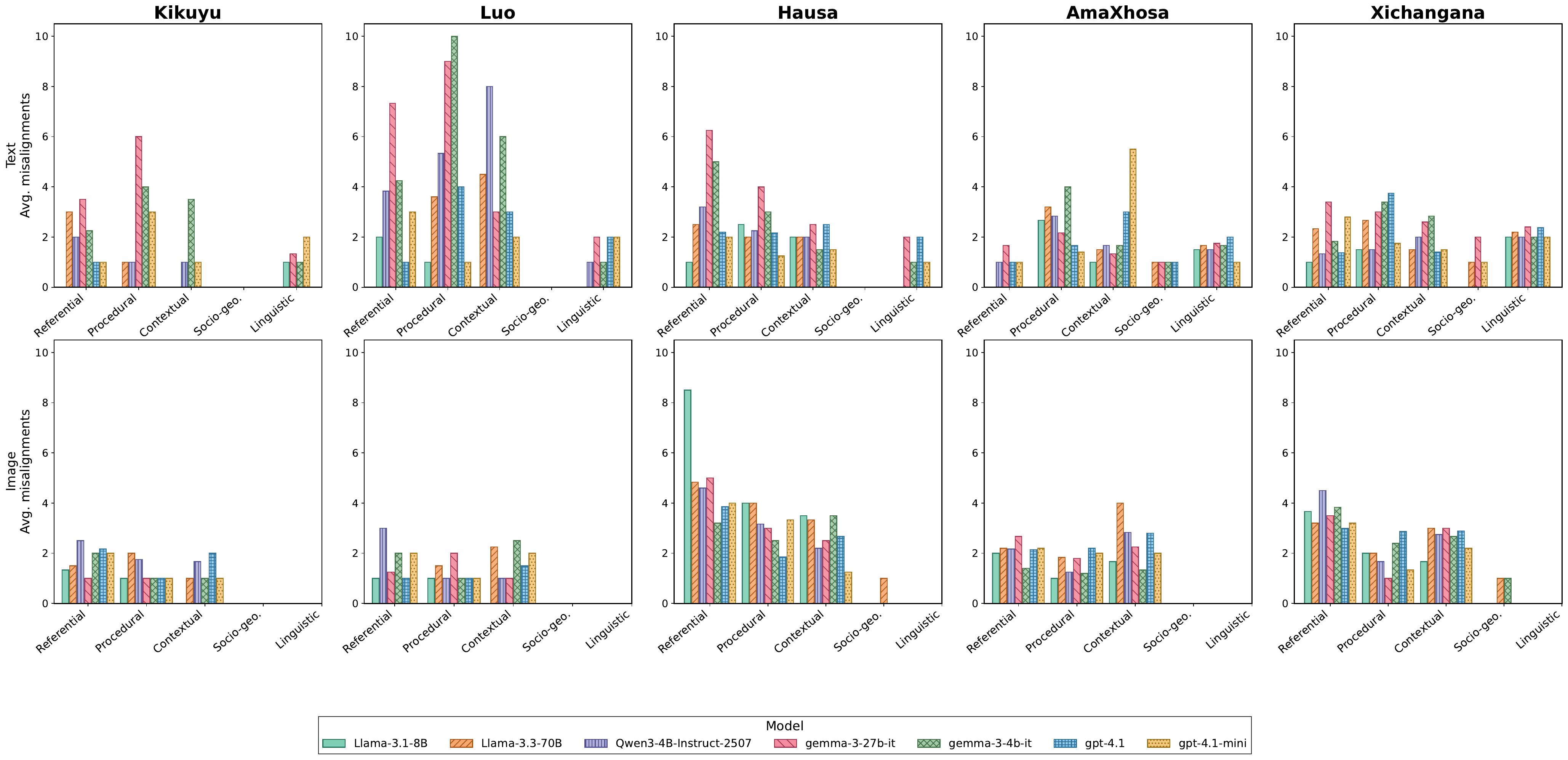}
    \caption{Distribution of not-connected cultural markers in text and images across the five broader cultural marker categories, broken down by generation model and community. Connected marker distributions for both modalities appear in Appendix Figure~\ref{fig:image-alignments-taxonomy}.}
    \label{fig:rq1}
\end{figure*}

\subsubsection{Cultural Marker Patterns Across Modalities}

Marker patterns varied across communities, modalities, and the five broader cultural marker categories. At the marker level, representatives labeled selected text spans and image regions as \textit{Connected} when they supported cultural fit and \textit{Not connected} when they weakened or disrupted cultural fit. Among text span annotations, 8,075 of 9,386 (86\%) were labeled connected and 1,311 (14\%) were labeled not connected. Image region annotations showed more visible disruption, with 6,931 of 9,419 (74\%) labeled connected and 2,488 (26\%) labeled not connected. Full text-span and image-region statistics are provided in Appendix Tables~\ref{tab:text-annotation-stats} and~\ref{tab:image-annotation-stats}. Figure~\ref{fig:rq1} descriptively shows the distribution of not-connected markers across communities, models, modalities, and cultural marker categories, while Appendix Figure~\ref{fig:image-alignments-taxonomy} shows the corresponding distribution of connected markers. These figures show that cultural fit and cultural disruption are not evenly distributed across modalities. In text, not-connected markers were often concentrated in different categories for different communities. Hausa stories showed frequent not-connected markers in \textit{Referential} and \textit{Procedural} categories, while Xichangana stories showed a more even spread across categories, including \textit{Linguistic Register}. In images, not-connected markers were often more visually concentrated, especially around character appearance, clothing, food, setting, public space, and other visual details. These patterns show that the visual realization of a story can support, weaken, transform, or contradict cultural cues in the narrative, making evaluation across both text and images necessary.

\subsubsection{Misalignment Mechanisms}

The focus group discussions revealed eight recurring mechanisms through which stories became culturally misaligned. \textbf{Substitution} occurred when stories included markers from another community in place of markers from the target community. For instance, \textit{sadza}, a Zimbabwean food, was independently flagged in three non-Zimbabwean communities: Kikuyu, Luo, and AmaXhosa. \textbf{Hallucination} occurred when models used cultural vocabulary but attached it to implausible practices. Hausa participants flagged an incorrect food preparation description, ``boil your tuwo shinkafa instead of frying,'' noting that frying was not a way they would cook this dish. An AmaXhosa participant similarly found \textit{umxhentso}, a traditional dance, described in a story as a food and commented, ``This is not an error a human storyteller would make. This is very AI.'' \textbf{Forced insertion} occurred when correct names or cultural terms appeared in otherwise generic narratives without meaningful integration, as when a single Kikuyu name was placed in a story where surrounding markers belonged to other cultures, making the name feel ``thrown in.''

\textbf{Norm violation} appeared when outputs contradicted expectations around respect, dress, or social interaction. A Xichangana participant explained, ``ninguém entra no hospital, participa de uma consulta de chapéu, considera-se como uma falta de respeito'' (nobody enters a hospital wearing a hat; it is considered a lack of respect). \textbf{Register conflation} was especially visible in Xichangana stories, where Portuguese was fluent but regionally inappropriate. One participant observed, ``usava-se muito o gerúndio e o gerúndio é muito característico dos brasileiros'' (the gerund was used a lot, and the gerund is very characteristic of Brazilians). \textbf{Cross-modal inconsistency} occurred when text and image represented conflicting settings, identities, or practices, while \textbf{stereotyping} appeared when models reduced communities to narrow or repeated visual tropes.

Finally, \textbf{omission} appeared when stories contained no obviously wrong elements but also few cultural details. Participants across communities described such stories as generic rather than factually incorrect. An AmaXhosa participant called one story ``a story anybody from any culture could narrate,'' a Luo participant described another as ``not grounded culturally,'' and a Xichangana participant said, ``não encontrei marcadores'' (I did not find markers). These mechanisms show that cultural misalignment is not only a matter of factual error, but also of weak integration, inappropriate context, cross-modal mismatch, wrong register, stereotyping, and absence of situated detail. See Appendix Figure~\ref{fig:cultural-alignment-taxonomy} for additional examples.

\subsection{LLM Judges Approximate Community Judgments Unevenly Across Cultures}
\label{sec:results-judge-validation}
\begin{table}[h!]
\centering
\scriptsize
\begin{tabular}{lccccc}
\toprule
Community & Kimi & Gemma4 & GPT-5.5 & Qwen-122B & Qwen-9B \\
\midrule
Hausa & \textbf{0.74} & \textbf{0.67} & \textbf{0.70} & \textbf{0.80} & \textbf{0.76} \\
AmaXhosa & 0.33 & 0.26 & 0.28 & 0.29 & 0.40 \\
Kikuyu & \textbf{0.54} & \textbf{0.50} & \textbf{0.51} & \textbf{0.64} & \textbf{0.57} \\
Luo & \textbf{0.89} & \textbf{0.82} & \textbf{0.88} & \textbf{0.87} & \textbf{0.87} \\
Xichangana & \textbf{0.49} & \textbf{0.45} & \textbf{0.51} & \textbf{0.49} & 0.39 \\
\bottomrule
\end{tabular}
\caption{Pearson correlation coefficients ($r$) between LLM judge scores and mean scores from culture representatives by community. Bold values remain significant after Bonferroni correction ($p<0.002$); unbolded values do not, even when significant at uncorrected thresholds ($p<0.001$, $p<0.01$, or $p<0.05$). No AmaXhosa judge remains significant after correction.}
\label{tab:judge-correlation}
\end{table}

Judge correlations with community scores varied across the five communities. For Luo, all five judges achieved very strong Pearson correlations with scores from culture representatives ($r=0.82$--$0.89$), and for Hausa correlations remained strong ($r=0.67$--$0.80$). Kikuyu and Xichangana fell in moderate ranges ($r=0.50$--$0.64$ and $r=0.39$--$0.51$, respectively). For AmaXhosa, no judge remained significant even after Bonferroni correction, indicating that automated scores did not reliably track community judgments for this community.

We first assess the reliability of the community ratings that serve as the reference point for judge validation and grounding analysis. Inter-annotator agreement on the overall cultural alignment score varied by community, ranging from good for Hausa (ICC(A,1)=0.81), to moderate for Luo, AmaXhosa, and Kikuyu (ICC(A,1)=0.60--0.63), to poor for Xichangana (ICC(A,1)=0.33). Xichangana's low agreement reflects a compressed score distribution (mean=24.5, SD=4.5), where representatives broadly rated stories as culturally inadequate but diverged on finer distinctions. All ICC values were statistically 
significant ($p<0.001$; Table~\ref{tab:icc}).

Score bias further limits automated evaluation. For Xichangana, judges assigned alignment scores 35--45 points higher than culture representatives (paired $t$-test, all $p<0.001$), assigning scores in the 60--69 range to stories that culture representatives rated at 24.5 on average. Kikuyu showed the opposite pattern, with Kimi and GPT-5.5 assigning scores 14--16 points lower than culture representatives. Judges were best calibrated on Hausa and showed smaller, mixed biases on Luo. Because both correlation with community scores and score bias varied by culture, no single judge model performed consistently across all five communities. Table~\ref{tab:judge-correlation} reports judge--community correlations, and Appendix Table~\ref{tab:judge-bias} reports score bias for all judge--community pairs.

\section{Discussion}
\label{sec:discussion}

Community evaluations show that cultural alignment is not reducible to the number or presence of recognizable markers. Culture representatives considered whether names, foods, places, clothing, language cues, social interactions, and visual settings fit the story context and the represented community. The taxonomy makes this distinction explicit by separating the broader cultural marker categories that representatives attended to from the mechanisms through which stories became misaligned. This matters for evaluation because a story may contain plausible markers while still failing through substitution, forced insertion, cross-modal inconsistency, wrong register, or absence of situated detail.

Our results also show that LLM judges do not approximate community judgments consistently across cultures. Judge reliability and score calibration varied substantially across communities, with no single judge performing consistently across all five settings. This variation means that automated judges should be treated as scalable evaluators only after community-specific validation. For cultural alignment tasks, both correlation with community judgments and score calibration should be considered, since a judge may track differences between stories while still systematically assigning scores that diverge from those of culture representatives.

These findings point toward community-calibrated evaluation pipelines for cultural alignment. In such pipelines, culture representatives provide the reference judgments needed to establish where automated evaluators can be trusted and where human review remains necessary. The taxonomy also makes community feedback more actionable for scalable evaluation: the five broader cultural marker categories identify what cultural dimensions automated judges should attend to, while the eight misalignment mechanisms describe recurring failure patterns that can guide review and calibration without replacing community judgment. Although the taxonomy was derived from five African communities in the context of diabetes lifestyle stories, its structure provides a basis for investigating whether similar cultural dimensions and misalignment mechanisms emerge in other communities and domains.

\section{Conclusion}
\label{sec:conclusion}

We presented a human-centered evaluation of cultural alignment in AI-generated multimodal stories across five African communities. Community evaluations showed that cultural alignment depends not only on the presence of recognizable cultural markers, but on how those markers fit narrative, social, linguistic, procedural, and visual context. From annotations and focus group discussions, we developed a taxonomy of five broader cultural marker categories and eight recurring mechanisms of misalignment. We further evaluated multimodal LLM judges against culture-representative judgments and found that their reliability and score calibration vary substantially across communities, with no single judge performing consistently across all five settings. These findings point toward community-calibrated evaluation pipelines in which community judgments ground cultural alignment evaluation and automated judges are validated to determine where they can be trusted and where human review remains necessary.

\section*{Limitations}
\label{sec:limitations}

Our study has several limitations that point to important directions for future work. First, our community evaluation relies on 19 culture representatives. These participants provide situated lived and linguistic knowledge, but they do not represent the full diversity of views within each community. Cultural alignment is not fixed or uniform, and perspectives may vary across age, gender, region, language use, religion, class, and rural-urban experience. Future work should build broader and more iterative community review processes that include more diverse participants and examine how cultural alignment judgments vary within communities.

Second, our evaluation focuses on story-level and marker-level cultural alignment rather than downstream effects on readers. We do not test whether culturally aligned stories improve understanding, trust, behavior-change outcomes, or lifestyle decision-making. Future work should connect cultural alignment evaluation to reader-facing outcomes, asking not only whether communities recognize a story as culturally coherent, but also whether such coherence changes how people interpret, trust, adapt, or use AI-generated content.

Third, our study is grounded in everyday Type II diabetes lifestyle and behavior-change stories. This domain was chosen because diet, physical activity, household routines, and care practices are culturally situated, but the cultural marker patterns, misalignment mechanisms, and judge behavior observed here may differ in other domains. Future work should therefore validate the framework across other kinds of generated content and application settings.

Fourth, cross-community comparisons should be interpreted carefully. Xichangana stories differed from the other communities in both language context and evaluation patterns. They were generated in Portuguese for the Mozambican context, received much lower community scores on average, showed lower agreement among representatives, and were substantially over-scored by LLM judges. These differences make Xichangana an important case for understanding failures of automated evaluation, but they also caution against interpreting score differences across communities as direct measures of cultural alignment difficulty. Future work should examine how language choice, regional context, and local norms shape cultural alignment judgments.

Finally, our LLM judge analysis covers five multimodal judges and a fixed evaluation prompt. Judge reliability may change with different models, prompts, rubrics, or calibration methods. Rather than seeking a universal automated judge for cultural alignment, future work should develop community-calibrated evaluation pipelines that identify where judge scores can be trusted, where they are biased, and where human review remains necessary. Future work can also test whether community-derived taxonomies improve scalable evaluation by incorporating examples of cultural marker categories and misalignment mechanisms into judge prompts, audit protocols, or human review workflows, while validating these approaches separately in each community context.

\section*{Ethical Considerations}
\label{sec:ethics}

The generated stories were evaluated with the participation of culture representatives from each respective community. It is important to note that the stories do not provide medical diagnoses, prescriptions, or recommendations related to medication. Instead, they focus exclusively on culturally contextualized lifestyle and behavior-change narratives, particularly around healthy eating habits, physical activity, and everyday practices associated with managing Type II diabetes. Cultural alignment should not be interpreted as evidence of medical correctness or safety, and health-related generated content should undergo appropriate medical review in addition to community-based cultural evaluation.

\bibliography{custom}

\clearpage
\appendix

\section{Communities and Culture Representatives}
\begin{figure}[h!]
\centering
\includegraphics[width=0.5\textwidth]{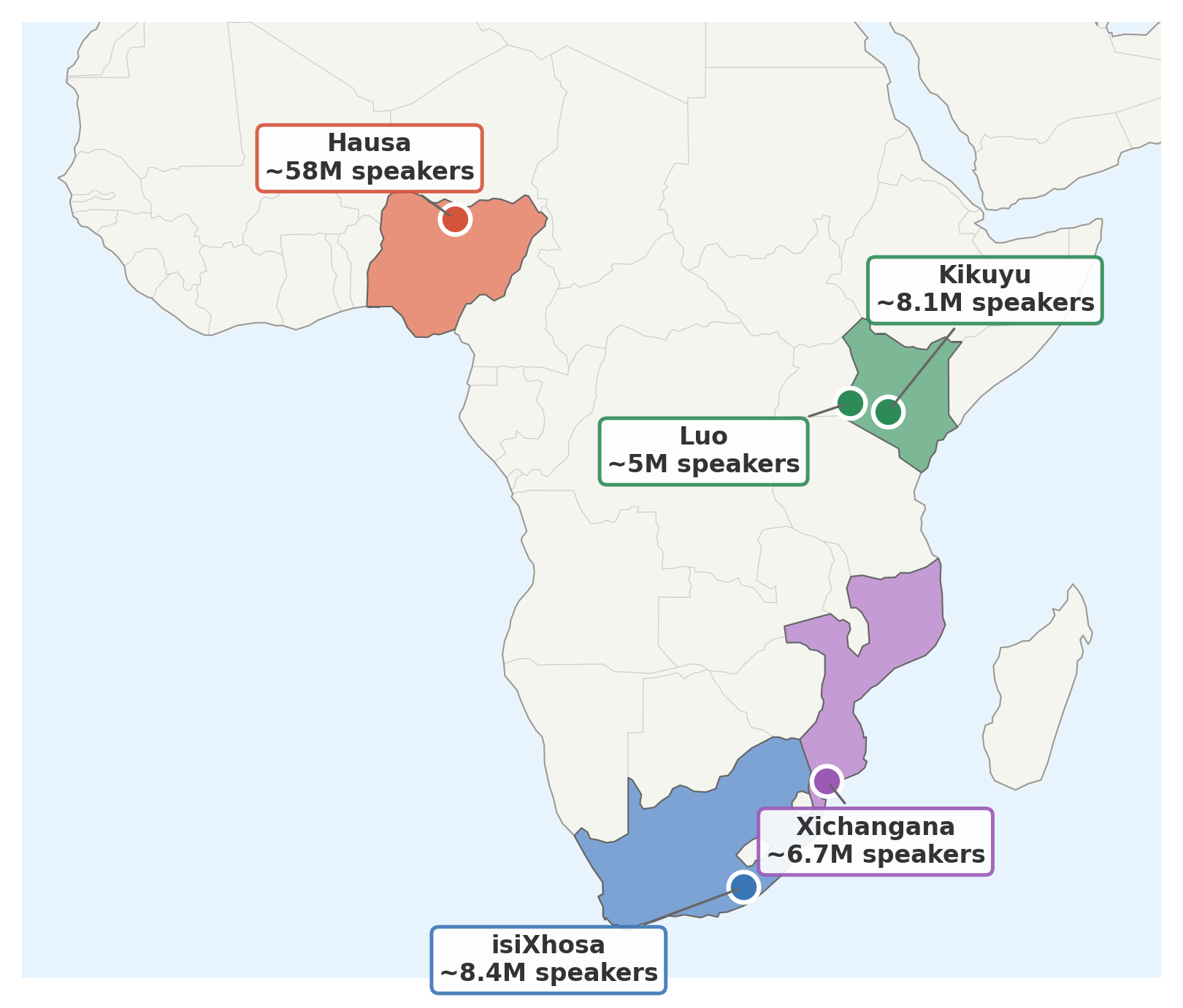}
\caption{Geographic distribution of the five African communities included in our study.}
\label{fig:cultures-map}
\end{figure}

Figure~\ref{fig:cultures-map} situates the five communities included in this study across the African continent, and Table~\ref{tab:culture-representative-demographics} summarizes participant demographics.
\label{app:communities}

Our study spans five African cultural communities across West, East, Southern, and Southeast Africa: Hausa in Nigeria, Kikuyu and Luo in Kenya, AmaXhosa in South Africa, and Xichangana in Mozambique. These communities differ in language, geography, religious and historical context, naming practices, foodways, dress norms, and everyday social expectations. We therefore treat them as distinct evaluative contexts.

The Hausa community in this study is situated in northern Nigeria, where Islamic practice, market life, Hausa naming conventions, modest dress expectations, and foods such as \textit{tuwo}, \textit{masa}, and \textit{miyan kuka} shape everyday cultural interpretation. The Kikuyu community is situated in central Kenya, where stories often draw on settings such as Kiambu, agricultural livelihoods, local foods, family relations, and Kikuyu naming practices. The Luo community is situated in western Kenya, where Luo names, kinship relations, lakeside and rural--urban settings, foods, and social forms of respect provide important cultural context. The AmaXhosa community is situated in South Africa, where AmaXhosa language practices, kinship relations, forms of address, dress, and institutional interactions shape how stories are interpreted. The Xichangana community is situated in Mozambique, where local Portuguese usage, Xichangana language and expressions, food practices, respect norms, and public behavior distinguish the community from both European Portuguese and other African contexts.

\begin{table}[h!]
\centering
\scriptsize
\begin{tabular}{lccc}
\toprule
\textbf{Community} & \textbf{n} & \textbf{Age Range} & \textbf{Gender} \\
\midrule
Hausa & 3 & 30--39 & 3F \\
Kikuyu & 4 & 18--34 & 3F, 2M \\
Luo & 6 & 18--44 & 4F, 2M \\
AmaXhosa & 2 & 25--29 & 2F \\
Xichangana & 4 & 18--44 & 3F, 1M \\
\midrule
Total & 19 & 18--44 & 15F, 5M \\
\bottomrule
\end{tabular}
\caption{Culture representative demographics by community.}
\label{tab:culture-representative-demographics}
\end{table}

\section{Story Generation Details}
\label{app:generation}

This appendix provides supplementary materials for the story generation pipeline described in Section~\ref{sec:storygeneration}, including the workflow, persona fields, lifestyle questions, story distribution, and narrative arcs.

\begin{table}[h!]
\scriptsize
\centering
\begin{tabular}{p{1cm} p{5.5cm}}
\toprule
\textbf{Category} & \textbf{Question} \\
\midrule
\multirow{5}{*}{Diet}
& Which everyday foods can I eat that will not make my blood sugar spike? \\
& I like to snack---what are some healthy options, and how often should I have them? \\
& I sometimes eat late at night. How does this affect my blood sugar, and what can I do instead? \\
& Can you give me some simple meal ideas or a sample daily menu I could follow? \\
& Eating healthy can get expensive---how can I maintain a nutritious diet on a tight budget? Can I still enjoy traditional foods? \\
\midrule
\multirow{5}{*}{Exercise}
& What kinds of exercises are safe and beneficial for someone with diabetes? \\
& Can I rely only on walking for exercise, or should I include other types of physical activity? \\
& How often and for how long should I exercise to improve my health? \\
& I do not have access to a gym---what activities can I do at home or with everyday items to stay active? \\
& Sometimes I feel tired or discouraged---how can I stay motivated, and how does exercise help control blood sugar? \\
\bottomrule
\end{tabular}
\caption{Everyday diabetes lifestyle questions used for narrative generation.}
\label{tab:health-questions}
\end{table}

\begin{table}[h!]
\centering
\scriptsize
\begin{tabular}{p{2.2cm}p{4.9cm}}
\toprule
\textbf{Context setting} & \textbf{Information provided during generation} \\
\midrule
Sparse persona & A reduced persona profile in which approximately half of the persona attributes were removed while preserving core demographic and cultural information. \\
Full persona & The complete generated persona profile. \\
Full persona + retrieval & The complete generated persona profile together with retrieved web context. GPT-4.1 generated search queries from the persona profile, and the retrieved information was provided as additional context for story generation. \\
\bottomrule
\end{tabular}
\caption{Context settings used to vary the amount of persona and cultural information available during story generation.}
\label{tab:context-settings}
\end{table}

\begin{table}[h!]
\centering
\scriptsize
\begin{tabular}{p{0.06\columnwidth} p{0.85\columnwidth}}
\toprule
\textbf{\#} & \textbf{Persona field} \\
\midrule
1 & Culture (select all that apply) \\
2 & Which province do you live in? (select 1) \\
3 & What is your religion? \\
4 & What is your preferred name? \\
5 & Gender (select 1) \\
6 & Age (select 1) \\
7 & How would you describe your local community or area? (select all that apply) \\
8 & How would you describe access to food in your local community or area? (select all that apply) \\
9 & Do you have a disability? (select all that apply) \\
10 & What is your current occupation? (select 1) \\
11 & What best describes your daily routine right now? \\
12 & What hobbies or interests do you enjoy regularly? (select all that apply) \\
13 & Which activities or routines are most important to you? \\
14 & Do you currently have other health conditions besides diabetes? (check all that apply) \\
15 & Are you currently taking medication or following a treatment for diabetes? \\
16 & What kind of healthcare resources do you regularly access or use? \\
17 & Do you have regular access to healthcare services? \\
18 & How often do you access healthcare services? (choose 1) \\
19 & Are you currently living with Type 2 Diabetes? \\
20 & Do you smoke tobacco products? \\
21 & If yes, how often? (choose 1) \\
22 & Do you consume alcoholic beverages? \\
23 & If yes, how often? (choose 1) \\
24 & How many days per week do you participate in moderate to vigorous physical activity? (choose 1) \\
25 & Select any dietary preferences or restrictions you may have (choose 1) \\
26 & Who forms part of your regular support network? (check all that apply) \\
27 & What are your main personal health goals? (You may select more than one) \\
28 & How do you prefer to receive information or stories about your health? (check all that apply) \\
\bottomrule
\end{tabular}
\caption{Persona fields used to construct culturally situated personas.}
\label{tab:profile-fields}
\end{table}

\onecolumn

\section{Annotation Interface and Evaluation Protocol}
\label{app:annotation}
This appendix provides the annotation platform, story-level scoring rubric, and focus group guiding questions referenced in Section~\ref{sec:methodology}.

\begin{table*}[t]
\centering
\small
\resizebox{\textwidth}{!}{%
\begin{tabular}{lll}
\toprule
\textbf{Category} & \textbf{Description} & \textbf{Examples} \\
\midrule

people\_appearance & skin tone, hair, age, who is present, visual identity cues & An elderly woman with grey hair greets a young boy at the gate. \\

names\_forms\_of\_address & forms of address that signal identity or relationships & “Auntie Mary” calls him “my child” when he arrives home. \\

language\_local\_expression & slang, code-switching, local phrases: “Ag”, “mos”, “lekker” & “Ag, don’t worry,” she says, switching between languages mid-sentence. \\

food\_dietary\_practices & foods, preparation methods, meal styles, eating habits & The family shares stiff porridge and vegetables for dinner. \\

place\_physical\_environment & province, home, clinics, kitchen, shops, neighborhoods, local setting & The small shop near the dusty road opens early every morning. \\

healthcare\_community\_practices & clinic visits, nurses, advice-seeking norms, care pathways & She visits the clinic after neighbors advise her to check her fever. \\

socio\_economic\_context & budgeting, shopping specials, affordability, access to food/resources & He counts coins carefully before buying bread for the week. \\

family\_household\_structure & family roles, caregiving, shared meals, household dynamics & Grandmother cooks while the children set the table. \\

occupation\_daily\_routine & jobs, lunch prep, meetings, time pressure, daily responsibilities & The teacher prepares lessons late at night after working all day. \\

health\_values\_beliefs & motivation, balance, “small tweaks” mindset, change & She decides to walk daily to “stay strong and healthy.” \\

clothing\_fashion & clothing, accessories, style, trends & He wears a school uniform neatly pressed for assembly. \\

social\_roles\_hierarchy & age, gender, seniority, authority, respect norms & The youngest waits silently while elders speak first. \\

community\_social\_networks & neighbors, mutual aid, social obligations, communal support & Neighbors come together to help rebuild a burned house. \\

gender\_norms\_expectations & gendered responsibilities, speech norms, social constraints & She is expected to cook while her brother studies. \\

life\_stage\_transitions & childhood, adulthood, parenting, aging-related roles & After graduating, he begins supporting his younger siblings. \\

temporal\_orientation & time-use norms, punctuality, event-based vs clock-based time & The meeting starts “after prayers,” not at a fixed hour. \\

seasonality\_environmental\_cycles & weather, harvest cycles, seasonal constraints & During the rainy season, roads become difficult to use. \\

daily\_rhythms\_meal\_patterns & when activities and meals occur during the day & Breakfast is early, before everyone leaves for work and school. \\

religious\_spiritual\_practices & faith traditions, rituals, moral frameworks & The family gathers for evening prayer before dinner. \\

traditional\_indigenous\_knowledge & local healing, food knowledge, environmental understanding & Her grandmother teaches her which leaves cure stomach aches. \\

health\_illness\_explanatory\_models & causes, symptoms, and recovery understanding & They believe the illness came from cold night air. \\

values\_moral\_frameworks & what is considered good, responsible, shameful, or desirable & “Helping others is more important than keeping everything for yourself.” \\

technology\_media\_use & phone sharing, platform preference, data constraints, media habits & They share one phone to check messages and news. \\

transport\_mobility & commuting methods, travel norms, access constraints & She takes a crowded minibus to work every morning. \\

household\_resources\_tools & appliances, cooking tools, storage, water/electricity access & The family cooks on a small charcoal stove outside. \\

communication\_style & directness, politeness strategies, humor, indirect speech & He hints politely instead of saying no directly. \\

emotional\_expression\_norms & what emotions are shown or hidden & She smiles in public but worries quietly at home. \\

storytelling\_narrative\_forms & personal stories, advice-giving, metaphor use & He tells a story about his childhood to give advice. \\

social\_norms\_taboo & topics avoided, prohibited behaviors, euphemisms & No one speaks openly about conflict during the family gathering. \\

institutional\_trust\_relations & attitudes toward government, clinics, NGOs, authority & They wait in doubt, unsure if the clinic will help. \\

silence\_implicit\_knowledge & assumed knowledge, omissions, things left unsaid & Everyone understands the problem without it being explained. \\

animals\_human\_interaction & livestock, pets, wildlife roles in livelihood & Goats roam freely near the homestead and are herded back at dusk. \\

plants\_agriculture\_food\_sources & crops, farming practices, gardens, wild foods & The family harvests maize from their small field. \\

sports\_recreation\_physical\_activity & organized sports, informal games, physical activity norms & Children play football barefoot in the open field. \\

arts\_entertainment\_culture & music, dance, festivals, performance & Drums and dancing fill the village during the celebration. \\

media\_information\_ecosystem & radio, TV, news, misinformation, access & The radio announces local news every evening. \\

built\_environment\_infrastructure & roads, sanitation, buildings, utilities & A dirt road connects the village to the town. \\

objects\_material\_culture & tools, artifacts, everyday items & A worn cooking pot sits over the fire. \\

symbols\_visual\_culture & flags, colors, religious symbols, logos & A painted symbol marks the entrance to the community hall. \\

events\_rituals\_celebrations & holidays, ceremonies, weddings, funerals & The wedding ceremony continues late into the night with music. \\

education\_learning\_practices & schooling, literacy, informal learning & The child practices reading aloud after school. \\

governance\_politics\_civic\_life & political structures, civic participation & The village leader calls a community meeting. \\

economy\_livelihood\_strategies & income sources, trade, markets, subsistence & She sells vegetables at the local market each morning. \\

environment\_ecology\_sustainability & resources, conservation, climate adaptation & Farmers adjust planting after noticing changing rainfall. \\

risk\_safety\_security & crime, hazards, coping mechanisms & They avoid walking home late due to safety concerns. \\

migration\_mobility\_patterns & rural-urban movement, diaspora, seasonal migration & He moved to the city to find work. \\

identity\_ethnicity\_culture & ethnic groups, heritage, identity markers & She proudly wears clothing that reflects her heritage. \\

disability\_accessibility & assistive tools, inclusion, barriers & The school adds ramps for wheelchair access. \\

food\_markets\_distribution & markets, vendors, supply chains & Fresh produce arrives early at the town market. \\

water\_sanitation\_hygiene & WASH practices, infrastructure, hygiene norms & They fetch water from a shared community tap. \\

energy\_fuel\_usage & cooking fuel, electricity sources & The household relies on firewood for cooking. \\

visual\_scene\_composition & camera angle, framing, indoor/outdoor cues & The photo shows a crowded street from above. \\

image\_artifacts\_quality & lighting, blur, resolution, filters & The image is blurry but still recognizable. \\

branding\_advertising\_presence & logos, billboards, sponsorship & A billboard advertises a popular drink along the road. \\

digital\_platform\_signals & UI elements, apps, emojis, memes & A WhatsApp message includes voice notes and emojis. \\

interpersonal\_distance\_body\_language & gesture, posture, touch norms & They stand close while talking, showing familiarity. \\

sound\_audio\_cues & music, noise, environmental sounds & Roosters crow as the village wakes up. \\
\textbf{other} & \textbf{if a category is missing, please provide} & \\

\bottomrule
\end{tabular}%
}
\caption{Cultural marker categories used for both text and image annotation. The same categories were applied across modalities, with annotators selecting text spans or image regions that influenced their cultural alignment judgments.}
\label{tab:marker-categories}
\end{table*}

\begin{table}[h!]
\scriptsize
\centering
\begin{tabular}{p{1.5cm} p{5.5cm}}
\toprule
\textbf{Score Range} & \textbf{Interpretation} \\
\midrule
0 & No recognizable cultural markers or cultural relevance. \\
1--20 & Minimal mention of cultural elements without meaningful integration. \\
21--40 & Some cultural markers are present, but integration is weak, generic, or only loosely relevant. \\
41--60 & Moderate cultural reflection with partial relevance to the represented community. \\
61--80 & Most cultural markers are meaningfully incorporated and broadly aligned with the represented community. \\
81--100 & Strong cultural alignment, with high relevance, coherence, and authenticity in relation to the represented community. \\
\bottomrule
\end{tabular}
\caption{Overall cultural alignment scoring rubric used by culture representatives and LLM judges.}
\label{tab:score-rubric}
\end{table}

\begin{figure*}[h]
    \centering
    \includegraphics[width=\textwidth]{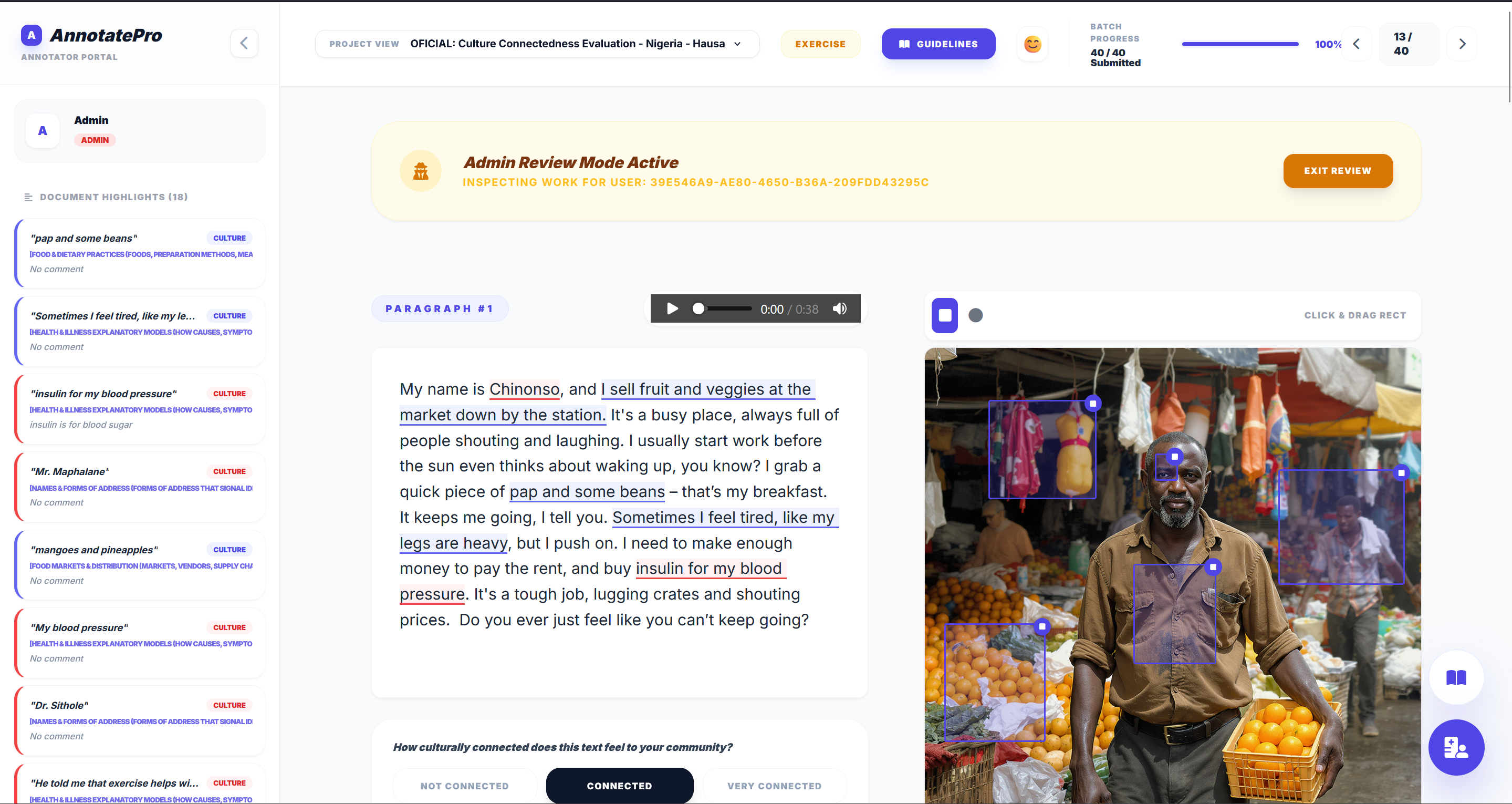}
    \caption{Annotation platform used by culture representatives to evaluate text and image cultural alignment.}
    \label{fig:platform1}
\end{figure*}

\begin{figure*}[h]
    \centering
    \includegraphics[width=0.8\textwidth]{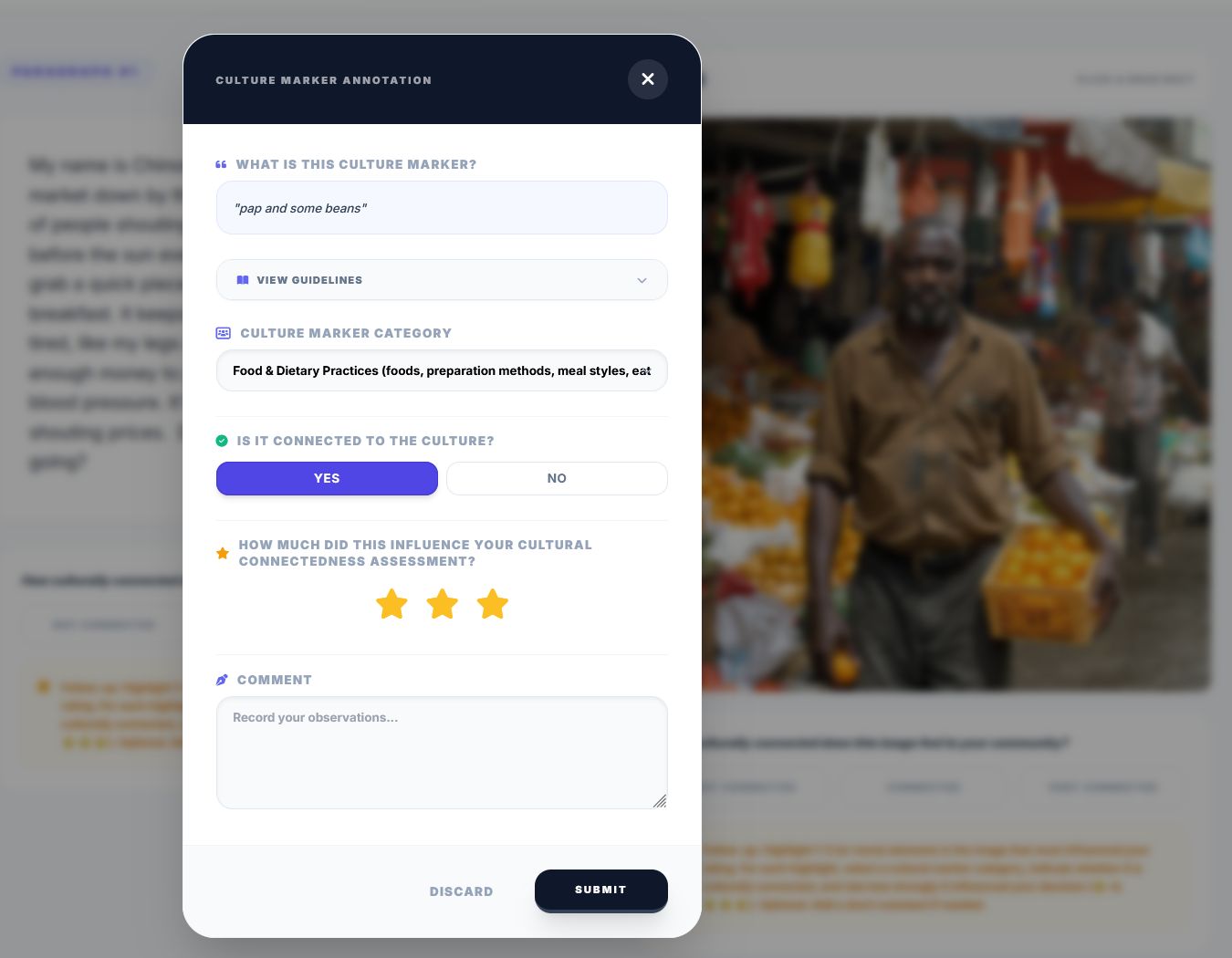}
    \caption{Annotation platform: Assigning a cultural marker category to a highlighted text span or region.}
    \label{fig:platform2}
\end{figure*}

\begin{figure*}[h]
    \centering
    \includegraphics[width=0.8\textwidth]{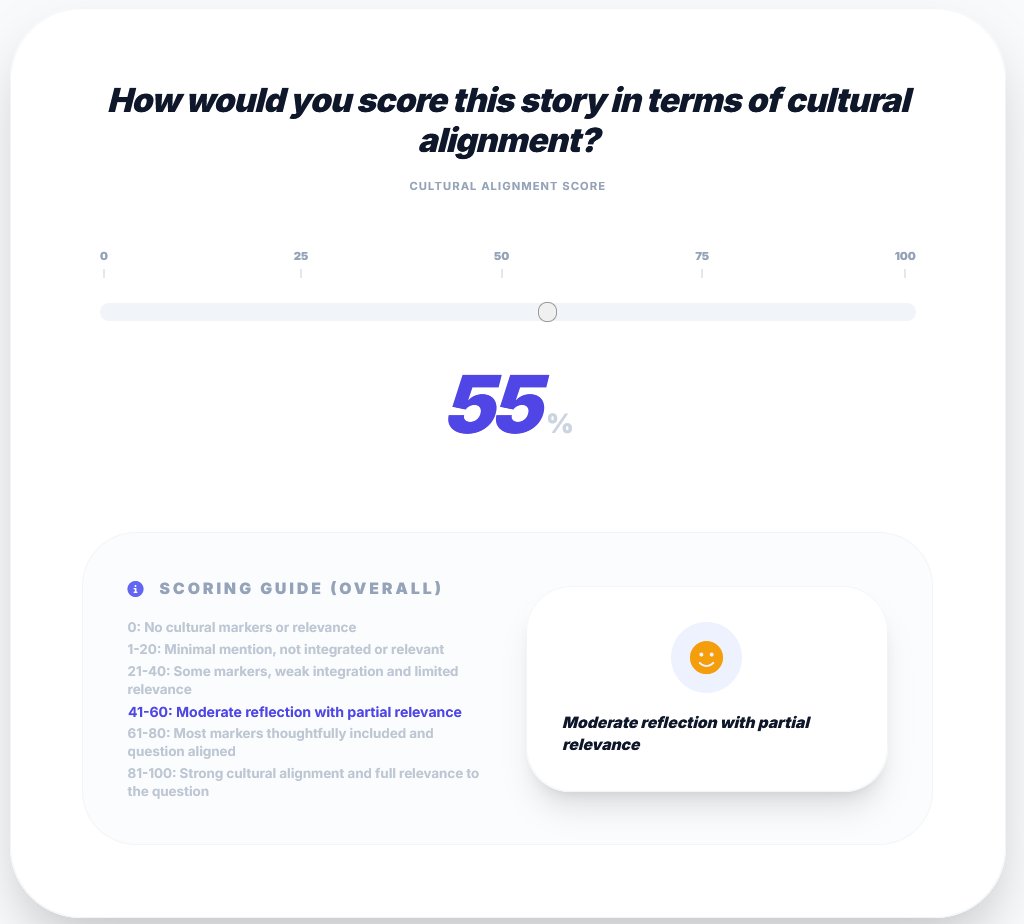}
    \caption{Annotation platform: Assigning overall culture alignment scoring.}
    \label{fig:platform3}
\end{figure*}

\clearpage
\begin{table*}[h]
\centering
\small
\begin{tabular}{p{3.2cm}p{12cm}}
\toprule
\textbf{Discussion focus} & \textbf{Guiding questions} \\
\midrule
Initial reactions &
\begin{enumerate}[leftmargin=*, nosep]
    \item After reading this story, did it feel like something that could realistically happen in your community? Why or why not?
    \item Would someone from your community recognize themselves in this story? What makes you say that?
    \item Did anything immediately stand out to you as either strongly connected to your community or clearly not connected? What was it?
\end{enumerate}
\\
\midrule
Cultural marker identification and misalignment &
\begin{enumerate}[leftmargin=*, nosep]
    \item Can you walk us through specific parts of the story that felt ``off'' or did not ring true for your community?
    \item For those moments, what exactly made them feel wrong? Were they incorrect, out of place, too generic, or based on assumptions?
    \item Were there important cultural elements that you expected to see but were missing? What were they?
    \item Across the stories you reviewed, are there particular types of cultural elements that the AI tends to get wrong or struggle with?
\end{enumerate}
\\
\midrule
Meaningful integration of cultural markers &
\begin{enumerate}[leftmargin=*, nosep]
    \item When cultural elements appeared in the story, such as food, language, or daily practices, did they feel like a natural part of the story, or did they feel added in? Why?
    \item Can you give an example of a cultural element that felt genuinely well integrated, and one that felt forced or out of place?
    \item Did some stories feel more grounded in your community than others? What specifically made the difference?
    \item In stories where cultural elements were present but the story still felt off, what was missing for it to feel truly connected to your community?
\end{enumerate}
\\
\bottomrule
\end{tabular}
\caption{Guiding questions used in focus group discussions with culture representatives.}
\label{tab:focus-group-protocol}
\end{table*}

\begin{table*}[h!]
\centering
\scriptsize
\begin{tabular}{lp{3cm}p{3cm}p{3cm}p{3cm}}
\toprule
 & Paragraph 1 & Paragraph 2 & Paragraph 3 & Paragraph 4 \\
\midrule
default &  Introduce \{px\_name\}’s current **daily life and diet habits** as described above. 
Mention any **cultural and environmental factors** that make change difficult.
The persona currently has a small work-related task that foreshadows a larger work-related task. They fail to accomplish the task because of issues related to health changes they need to make. All motivation to change comes from the need to accomplish the current and future task.
 &  Because of the failed small task, \{px\_name\} needs help. Depending on the given healthcare access \& resources, either \{px\_name\} seeks help e.g., visits a **clinic or community health worker**, or help finds them coincidentally e.g., support network. This helper gives practical advice on managing their lifestyle or current challenge. The persona is initially resistant to the change. Motivation to change comes from failure at the small task. The health worker suggests a small change that will help the persona accomplish the task.
 &  Describe **how \{px\_name\} tries to follow the advice given**, facing initial struggles due to certain reason(s). Describe how they overcome these struggles. 
The persona tries to make the prescribed health change but faces a setback. They try to find alternative ways to pursue the health change in light of the upcoming larger task. 
 &  ** Highlight **positive but realistic changes**, that result from the new effort they've put in.
The persona finds the small changes helpful in health management. This enables them accomplish the larger task. The persona recognises the health AND work-related benefits of the health changes. The persona is motivated to make more / bigger changes to their health habits.  \\ \midrule

classic change &   Introduce daily life and diet habits that reflect **limited awareness** of how health affects work. The small work task fails due to misunderstanding or neglect of health needs. &  Help comes from a **formal health source** (clinic or health worker). Resistance is mild and based on uncertainty. Advice is clear, simple, personalized, culturally relevant, actionable, and practical. &   The persona attempts the advice but struggles with **adjusting routines**. A minor setback occurs due to habit change. They persist because of the upcoming larger task. &    Gradual, realistic improvement occurs. The larger task is accomplished with more confidence. The persona becomes open to continued health improvement. \\\midrule

crisis avoidance &   Daily habits show warning signs or discomfort.
  Failure of the small task includes a **health scare or limitation**. &    Help comes from a trusted health authority. Advice (clear, simple, personalized, culturally relevant, actionable, and practical) emphasizes preventing serious future consequences. Resistance is rooted in fear or denial. &   The persona attempts the change but hesitates or backslides due to anxiety. They reframe the change as necessary to succeed at the larger task. &   Health stabilizes in a realistic way. The larger task is completed with relief. The persona is motivated to prevent future crises. \\\midrule
  
iteration &   The small task fails as part of **trial and error**, not a single mistake. Health issues are one of several contributing factors. &    Help provides practical advice, but it is not positioned as a perfect solution. Resistance is low; expectations are cautious. & The initial strategy does not work. A setback occurs. The persona modifies the advice (clear, simple, personalized, culturally relevant, actionable, and practical) creatively to fit their reality and the larger task.
 &   Learning from iteration leads to success in the larger task. Health improvements are modest but meaningful. The persona commits to ongoing adjustment rather than perfection. \\\midrule
resistance &   Introduce daily habits that are deeply ingrained. The small work task fails due to **stubborn routines or skepticism** about health change. & Help is offered by a health worker or trusted person. The persona doubts the advice (clear, simple, personalized, culturally relevant, actionable, and practical) and resists change. They agree to a small adjustment only to fix the work issue. &   The persona partially ignores or delays the advice. This leads to a setback. They reluctantly adjust their approach because the larger task is approaching. &   Resistance softens. The persona acknowledges both health and work benefits. \\ \midrule

social support &   Daily life is shaped by **social and cultural expectations**. The small work task fails due to competing social demands affecting health. &  Help comes from a **family member, coworker, or community figure**. Initial resistance comes from fear of disrupting social norms. Advice (clear, simple, personalized, culturally relevant, actionable, and practical) is framed as shared or collective.
 &   The persona tries to change but faces setbacks due to social situations. They adapt by involving others or adjusting timing. &   Positive changes are reinforced socially. The larger task is completed with support. The persona feels motivated to sustain change within their community. \\
\bottomrule
\end{tabular}
\caption{Story Arcs narrative progession}
\label{tab:story-arcs}
\end{table*}

\begin{landscape}

\begin{figure*}[h!]
    \centering
    \includegraphics[width=1\textwidth]{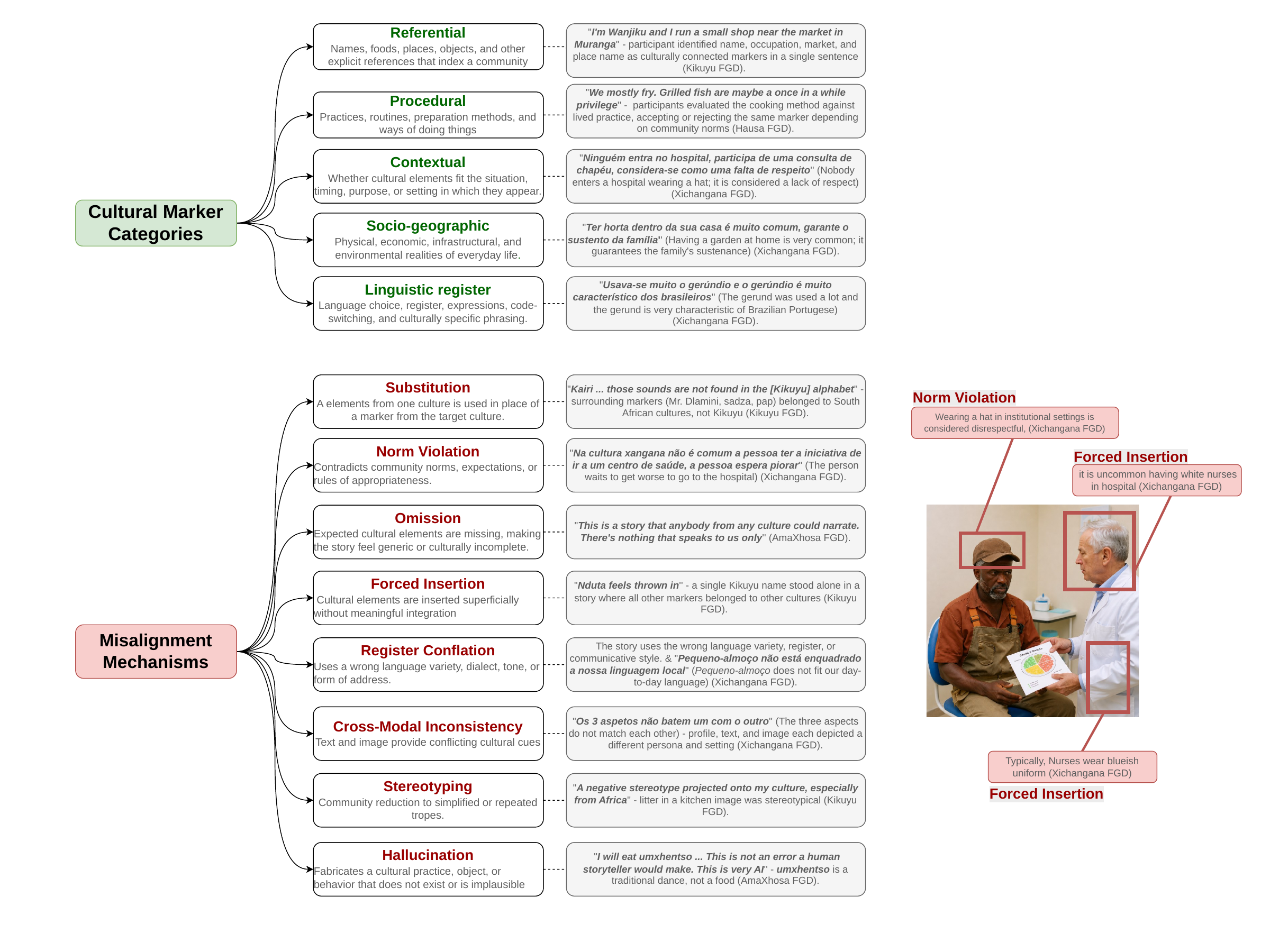}
    \caption{Cultural alignment taxonomy with examples from FGD sessions.}
    \label{fig:cultural-alignment-taxonomy}
\end{figure*}
\end{landscape}



\clearpage
\section{Prompts}
\label{app:prompts}

\begin{promptbox}{Generation System Prompt}
\scriptsize
\refstepcounter{figure}
\label{prompt:system}
\vspace{0.5em}

\textbf{Role:} A helpful hospital software operating in the patient’s location.

\textbf{Task:} Receive instructions and generate a first-person short story based in the patient’s location with a happy ending.

\vspace{0.5em}
The foods, clothing, and housing should reflect the financial status of neighbourhoods found in the patient’s location.

The narrative must:
\begin{itemize}[itemsep=0pt, topsep=0pt]
    \item Use first-person voice
    \item Sound like the person is directly telling their story
    \item Be culturally and contextually grounded
\end{itemize}

\vspace{0.5em}
\textbf{Constraints:}
\begin{itemize}[itemsep=0pt, topsep=0pt]
    \item For image prompts: \{sys\_img\_format\_flux\}
\end{itemize}

\vspace{0.5em}
\textbf{Output Requirements:}
\begin{itemize}[itemsep=0pt, topsep=0pt]
    \item Write the paragraphs first, story should first person account and all paragraphs must be in \{language\}.
    \item Then provide image prompts
    \item The image prompts should be written entirely in English.
    \item Return everything in JSON format
    \item Return only paragraphs written entirely in \{language\}.
    \item Do not include explanations or any other language.
\end{itemize}

\textbf{JSON Format:}
\begin{verbatim}
{sys_json_format}
\end{verbatim}

\end{promptbox}

\begin{promptbox}{Generation User Prompt}
\scriptsize
\refstepcounter{figure}
\label{prompt:user}

\vspace{0.5em}

Our persona has the following profile: \{persona\_profile\}

\vspace{0.5em}
Using the profile of \{px\_name\} and understanding the persona's needs ("\{question\}"), generate a four-paragraph health story written in \{language\} that centers around the needs of the persona, which is based on the following question: \{question\}. 

\vspace{0.5em}
The story will have a small work task that foreshadows a larger work task. This will serve as external motivation for the persona to make health changes.

\vspace{0.5em}
\{story\_arc\}

\end{promptbox}

\begin{promptbox}{Judge System Prompt}
\scriptsize
\refstepcounter{figure}
\label{prompt:llm-as-judge}

You are a cultural-alignment judge. Rate how culturally aligned an LLM-generated story is with its target culture. Be precise, cite text spans, and ground every claim in observable cultural markers. \\

\#\# Rating scale (0–100) \\
\{ 0: "No cultural markers or relevance", 1–20: "Minimal mention, not integrated or relevant", 21–40: "Some markers, weak integration and limited relevance", 41–60: "Moderate reflection with partial relevance", 61–80: "Most markers thoughtfully included and question-aligned", 81–100: "Strong cultural alignment and full relevance to the question"\} \\

\#\# Cultural marker categories (use these keys exactly) \\
\{\_MARKER\_LIST\} \\

\#\# Your task

\begin{enumerate}[itemsep=0pt, topsep=0pt]
    \item Read the story and original question.
    \item If paragraph-image pairs are provided, evaluate whether Paragraph 1 matches Image 1, Paragraph 2 matches Image 2, and so on.
    \item Check whether the story maintains cultural coherence across all paragraphs, not just isolated details.
    \item If persona or image context is provided, use it as supporting context only.
    \item For EACH cultural marker you find, quote the exact text span and assign it, a category key from the list above.
    \item For EACH important image-based cultural alignment/misalignment (using persona's context) or mismatch, add an item to Image\_cultural\_markers\_found and assign it, a category key from the list above and provide a comment
    \item  Assess depth and authenticity of cultural integration, including paragraph-image alignment.
    \item Penalize mismatched, generic, or culturally contradictory paragraph-image pairs.
    \item Score 0-100 and cite the bracket.
\end{enumerate}

\vspace{0.5em}
\#\# Output — respond with ONLY this JSON object, nothing else

\vspace{0.5em}
<think>

\end{promptbox}

\begin{promptbox}{Judge User Prompt}
\scriptsize
\refstepcounter{figure}
\label{prompt:judge-user}

\vspace{0.5em}

\#\# Culture \\
\{culture\} \\

\#\# Original question / prompt \\
\{question\} \\

\#\# Story \\
\{story\_text\} \\

\#\# Persona / profile \\
\{persona\} \\

\textit{
Assume Pair 1 maps to Paragraph 1 and Image 1, Pair 2 maps to Paragraph 2 and Image 2, and so on.
}

\vspace{0.6em}

\begin{tabular}{p{0.24\linewidth} p{0.70\linewidth}}
\textbf{Paragraph [idx + 1]} &
\texttt{\{paragraphs[idx]\}} \\[1em]

\textbf{Image [idx + 1] reference} &
\texttt{\{image\_urls[idx]\}}
\end{tabular}

\end{promptbox}


\newpage
\section{Additional Quantitative Results}
\label{app:additional-results}

\begin{table}[h!]
\centering
\scriptsize
\begin{tabular}{lccccc}
\toprule
Community & Kimi & Gemma4 & GPT-5.5 & Qwen-122B & Qwen-9B \\
\midrule
Hausa      & +1   & +7   & +3   & \textbf{+7}   & \textbf{+6} \\
AmaXhosa   & -1   & +4   & 0    & \textbf{+15}  & \textbf{+12} \\
Kikuyu     & \textbf{-16} & -7 & \textbf{-15} & -3 & -2 \\
Luo        & \textbf{-6}  & +1   & \textbf{-4}  & \textbf{+6} & +4 \\
Xichangana & \textbf{+38} & \textbf{+44} & \textbf{+36} & \textbf{+45} & \textbf{+45} \\
\bottomrule
\end{tabular}
\caption{Score bias, computed as judge mean minus culture-representative mean, by community. Bold values indicate significant paired $t$-test differences between judge and human scores. Xichangana inflation is systematic across all judges.}
\label{tab:judge-bias}
\end{table}

\begin{table}[h]
\centering
\scriptsize
\begin{tabular}{lcccc}
\toprule
Community & \#Annotators & ICC(A,1) & ICC(A,k) & Mean Score \\
\midrule
Hausa & 3 & 0.81 & 0.93 & 57.4 \\
AmaXhosa & 2 & 0.63 & 0.77 & 62.8 \\
Kikuyu & 4 & 0.60 & 0.86 & 76.2 \\
Luo & 6 & 0.63 & 0.91 & 65.6 \\
Xichangana & 4 & 0.33 & 0.66 & 24.5 \\
\bottomrule
\end{tabular}
\caption{Human annotator agreement per community. ICC(A,1) = absolute agreement for a single rater. ICC(A,k) = absolute agreement averaged across $k$ raters.}
\label{tab:icc}
\end{table}

\begin{figure*}[h!]
    \centering
    \includegraphics[width=1\textwidth]{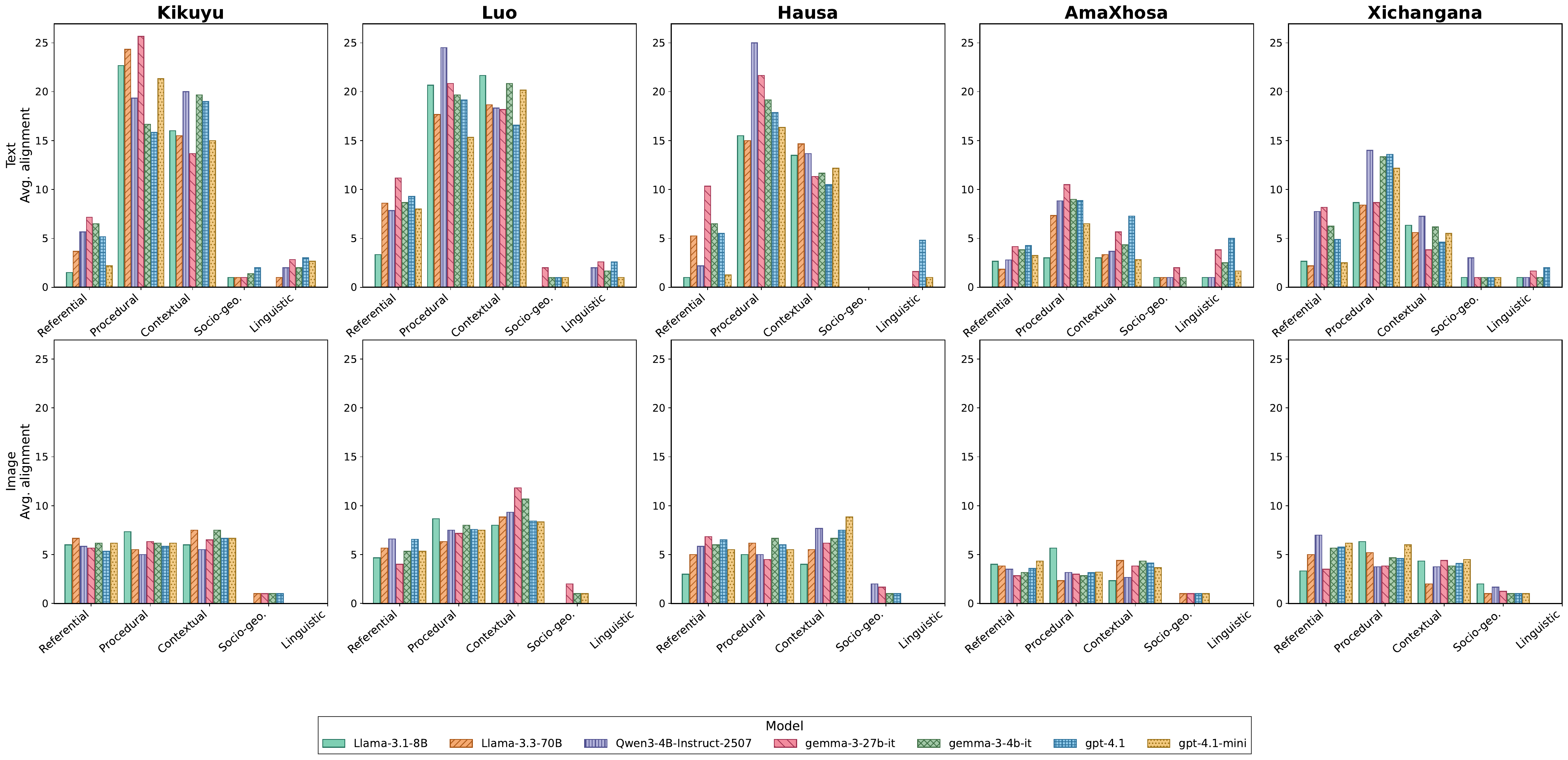}
    \caption{Distribution of connected cultural markers in text and images across the five broader cultural marker categories, broken down by generation model and community. This figure complements Figure~\ref{fig:rq1}, which shows not-connected markers.}
    \label{fig:image-alignments-taxonomy}
\end{figure*}

\subsection{Cultural Markers}
\label{sec:span-evaluation-results}

\paragraph{Data collected}
Culture representatives produced 18,805 span-level annotations across 199 stories
(9,386 text spans and 9,419 image regions). Of the text spans, 8,075 (86\%) were
labeled as culturally connected and 1,311 (14\%) as not connected. For image regions, 6,931 (74\%) were connected and 2,488 (26\%) not connected (Table~\ref{tab:text-annotation-stats}). These raw counts include overlapping annotations where multiple representatives
independently identified the same cultural marker. After deduplication (exact text
match for spans; Intersection over Union $>$ 0.3 clustering for image bounding boxes), the 8,075 connected text annotations correspond to 6,056 unique spans, of which 1,016 (16.8\%) were independently identified by more than one representative. Similarly, the 6,931 connected image annotations correspond to 5,435 unique regions, with 1,171 (21.5\%) identified by multiple representatives. The higher image overlap rate (21.5\% vs.\ 16.8\% for text) suggests that visually salient cultural markers in images are more consistently recognized across annotators than textual ones.

\begin{table*}[t]
\centering

\resizebox{\textwidth}{!}{%
\begin{tabular}{l c c r rrr rrr r rrr rrr}
\toprule
& & & & \multicolumn{3}{c}{\textbf{Conn. — Exact}} & \multicolumn{3}{c}{\textbf{Conn. — Partial}} & & \multicolumn{3}{c}{\textbf{Not C. — Exact}} & \multicolumn{3}{c}{\textbf{Not C. — Partial}} \\
\cmidrule(lr){5-7} \cmidrule(lr){8-10} \cmidrule(lr){12-14} \cmidrule(lr){15-17}
\textbf{Community} & \textbf{\# Ann.} & \textbf{\# Stories} & \textbf{Tot.} & \textbf{Uniq.} & \textbf{Ovlp.} & \textbf{\%} & \textbf{Uniq.} & \textbf{Ovlp.} & \textbf{\%} & \textbf{Tot.} & \textbf{Uniq.} & \textbf{Ovlp.} & \textbf{\%} & \textbf{Uniq.} & \textbf{Ovlp.} & \textbf{\%} \\
\midrule
Hausa      & 3 & 40 & 1{,}684 & 1{,}340 & 178 & 13.3 & 1{,}049 & 321 & 30.6 & 291 & 169 & 54 & 32.0 & 151 & 59 & 39.1 \\
AmaXhosa   & 2 & 40 & 781 & 674 & 95 & 14.1 & 644 & 103 & 16.0 & 207 & 191 & 6 & 3.1 & 187 & 9 & 4.8 \\
Kikuyu     & 4 & 39 & 2{,}068 & 1{,}570 & 263 & 16.8 & 1{,}323 & 345 & 26.1 & 94 & 65 & 18 & 27.7 & 56 & 18 & 32.1 \\
Luo        & 6 & 40 & 2{,}453 & 1{,}674 & 331 & 19.8 & 1{,}277 & 416 & 32.6 & 291 & 183 & 45 & 24.6 & 157 & 46 & 29.3 \\
Xichangana & 4 & 40 & 1{,}089 & 798 & 149 & 18.7 & 682 & 172 & 25.2 & 428 & 329 & 58 & 17.6 & 376 & 23 & 6.1 \\
\midrule
\textbf{Total} & 19 & 199 & 8{,}075 & 6{,}056 & 1{,}016 & 16.8 & 4{,}975 & 1{,}357 & 27.3 & 1{,}311 & 937 & 181 & 19.3 & 927 & 155 & 16.7 \\
\bottomrule
\end{tabular}%
}

\caption{Text span annotation statistics per language. \textit{Connected}: spans marked as culturally connected. \textit{Not connected}: spans marked as not culturally connected. Deduplication is shown at two levels: \textit{Exact} (identical span text within a task) and \textit{Partial} (token overlap $>$50\% of the shorter span with same category). \textit{Overlap}: spans independently identified by more than one representative.}
\label{tab:text-annotation-stats}

\end{table*}

\begin{table}[t]
\centering
\scriptsize
\begin{tabular}{l c c rrrr rrrr}
\toprule
& & & \multicolumn{4}{c}{\textbf{Connected}} & \multicolumn{4}{c}{\textbf{Not Connected}} \\
\cmidrule(lr){4-7} \cmidrule(lr){8-11}
\textbf{Community} & \textbf{\#Ann.} & \textbf{\# Stories} & \textbf{Tot.} & \textbf{Uniq.} & \textbf{Ovlp.} & \textbf{\%} & \textbf{Tot.} & \textbf{Uniq.} & \textbf{Ovlp.} & \textbf{\%} \\
\midrule
Hausa      & 3 & 40 & 1{,}854 & 1{,}476 & 323 & 21.9 & 960 & 773 & 147 & 19.0 \\
AmaXhosa   & 2 & 40 & 559 & 509 & 50 & 9.8 & 430 & 410 & 20 & 4.9 \\
Kikuyu     & 4 & 39 & 1{,}246 & 1{,}033 & 184 & 17.8 & 219 & 205 & 13 & 6.3 \\
Luo        & 6 & 40 & 2{,}141 & 1{,}476 & 447 & 30.3 & 324 & 272 & 40 & 14.7 \\
Xichangana & 4 & 40 & 1{,}131 & 941 & 167 & 17.7 & 555 & 460 & 70 & 15.2 \\
\midrule
\textbf{Total} & 19 & 199 & 6{,}931 & 5{,}435 & 1{,}171 & 21.5 & 2{,}488 & 2{,}120 & 290 & 13.7 \\
\bottomrule
\end{tabular}
\caption{Image region annotation statistics per language. Unique regions are identified by clustering bounding boxes with IoU (Intersection over Union) $>$ 0.3 across annotators. \textit{Overlap}: regions independently identified by more than one representative.}
\label{tab:image-annotation-stats}
\end{table}

\paragraph{Cultural Marker Span Evaluation}
To assess how well the LLM judge identifies culturally relevant text spans to support their reasoning, we adapted evaluation methodology from Named Entity Recognition (NER). This analysis focuses exclusively on the text modality; 
We constructed ground truth from human annotations where evaluators marked spans as culturally connected, aggregated across annotators via majority vote. We then evaluated the LLM judge predictions against this ground truth using three matching schemes: \textbf{Strict} (exact span text and category match), \textbf{Partial} (token overlap $>$50\% of the shorter span with correct category), and \textbf{Type} (correct category regardless of span boundaries).

Following NER evaluation literature, we adopt a proportional overlap threshold rather than the binary any-overlap criterion. The CoNLL shared tasks established exact-match as the standard for NER evaluation~\cite{tjong-kim-sang-de-meulder-2003-introduction}, while the MUC evaluation framework introduced partial matching categories (correct, incorrect, partial, missing, spurious) to capture boundary errors more granularly~\cite{chinchor-sundheim-1993-muc5}. SemEval-2013 further formalized four evaluation modes---strict, exact, partial, and type---where partial matching counts any span overlap as a match~\cite{segura-bedmar-etal-2013-semeval}. However, the any-overlap criterion assumes that predicted and gold spans have comparable granularity. In our setting, the LLM judge produces longer spans (mean 6.9 tokens, median 8) while human annotators mark shorter, phrase-level spans (mean 5.8 tokens, median 3), creating a structural asymmetry where coincidental single-token overlaps (e.g., function words such as ``the'', ``my'', ``I'') generate spurious matches between semantically unrelated spans. We therefore require $>$50\% token overlap relative to the shorter span, ensuring that matched pairs share substantive lexical content.

\subsubsection{Ground Truth Example}

Figure~\ref{fig:annotation-example} illustrates a ground truth annotation from a randomly selected Hausa story. Culture representatives identified specific text spans and assigned each to a cultural marker category. The example shows how a single story paragraph may contain overlapping cultural dimensions, such as names, dietary practices, local expressions, and occupation routines, that evaluators disambiguate at the span level.

\begin{figure}[h]
\centering
\small
\fbox{\parbox{0.92\textwidth}{%
 ``Hai, my name is \colorbox{green!20}{\textbf{Amina Yusuf}}\textsuperscript{\tiny names\_forms\_of\_address}, and life as a \colorbox{cyan!20}{\textbf{trader}}\textsuperscript{\tiny occupation\_daily\_routine} in \colorbox{blue!15}{\textbf{Kano}}\textsuperscript{\tiny place\_physical\_environment} market is\ldots busy, \colorbox{teal!20}{\textbf{wallahi}}\textsuperscript{\tiny language\_local\_expression}! Every day is a hustle. \colorbox{yellow!20}{\textbf{I sell beautiful fabrics, Ankara and lace}}\textsuperscript{\tiny economy\_livelihood\_strategies}, you know? Bright colours, good quality. But let me tell you, thinking about what to eat is always last on my mind. Usually, it's whatever's quickest -- maybe some \colorbox{red!20}{\textbf{masa}}\textsuperscript{\tiny food\_dietary\_practices} with a little \colorbox{red!20}{\textbf{stew}}\textsuperscript{\tiny food\_dietary\_practices}, or \colorbox{red!20}{\textbf{tuwo shinkafa}}\textsuperscript{\tiny food\_dietary\_practices} with \colorbox{red!20}{\textbf{miyan kuka}}\textsuperscript{\tiny food\_dietary\_practices}\ldots''
\vspace{4pt}\\

}}
\caption{Example ground truth annotation from the Hausa showing cultural marker spans identified by culture representatives. Each highlighted span is assigned a cultural category. The ground truth aggregates all connected spans across annotators via majority vote.}
\label{fig:annotation-example}
\end{figure}

We first measured pairwise agreement among human culture representatives to establish a ceiling for automated evaluation. Table~\ref{tab:human-iaa} reports average pairwise span-level F1 and category agreement across annotator pairs.

\begin{table}[h]
\centering

\begin{tabular}{lcc}
\toprule
\textbf{Language} & \textbf{Span F1} & \textbf{Category Agreement} \\
\midrule
Hausa      & 0.474 & 0.735 \\
AmaXhosa   & 0.420 & 0.646 \\
Kikuyu     & 0.447 & 0.678 \\
Luo        & 0.418 & 0.590 \\
Xichangana & 0.360 & 0.723 \\
\midrule
\textbf{Average} & \textbf{0.424} & \textbf{0.674} \\
\bottomrule
\end{tabular}
\caption{Human inter-annotator agreement on cultural marker spans (pairwise average). Only spans marked as culturally connected are included.}\label{tab:human-iaa}
\end{table}

Human annotators achieve moderate span overlap (F1 = 0.36--0.47) and moderate-to-substantial category agreement (0.59--0.74). This reflects the inherent subjectivity of cultural marker identification: annotators often agree on which cultural categories are present but differ in where they draw span boundaries. These scores establish an empirical upper bound for what can be expected from automated systems.

\subsubsection{LLM Judge Self-Consistency}

We ran the LLM judge three times under identical configurations and measured pairwise agreement across runs. Table~\ref{tab:model-consistency} reports inter-run agreement.

\begin{table}[h]
\centering

\begin{tabular}{lcc}
\toprule
\textbf{Language} & \textbf{Span F1} & \textbf{Category Agreement} \\
\midrule
Hausa      & 0.819 & 0.908 \\
AmaXhosa   & 0.813 & 0.909 \\
Kikuyu     & 0.793 & 0.914 \\
Luo        & 0.806 & 0.887 \\
Xichangana & 0.679 & 0.757 \\
\midrule
\textbf{Average} & \textbf{0.782} & \textbf{0.875} \\
\bottomrule
\end{tabular}
\caption{LLM judge inter-run agreement (pairwise average across 3 runs).}
\label{tab:model-consistency}
\end{table}

The model demonstrates high self-consistency (Span F1 = 0.68--0.82, Category Agreement = 0.76--0.91), substantially exceeding human agreement. This suggests that the judge produces stable outputs across repeated evaluations, though lower consistency for Xichangana suggests greater uncertainty for that language.

\subsubsection{LLM Judge vs.\ Human Ground Truth}

Table~\ref{tab:model-vs-gt} presents the main evaluation, LLM judge predictions, obtained by aggregating annotated text spans across judges through majority voting, compared against the human-annotated ground truth using NER-style metrics. Results are averaged across three runs.

\begin{table}[h]
\centering

\begin{tabular}{l ccc ccc ccc}
\toprule
& \multicolumn{3}{c}{\textbf{Strict}} & \multicolumn{3}{c}{\textbf{Partial}} & \multicolumn{3}{c}{\textbf{Type}} \\
\cmidrule(lr){2-4} \cmidrule(lr){5-7} \cmidrule(lr){8-10}
\textbf{Language} & P & R & F1 & P & R & F1 & P & R & F1 \\
\midrule
Hausa      & .080 & .080 & .080 & .196 & .197 & .196 & .459 & .461 & .460 \\
AmaXhosa   & .054 & .110 & .072 & .129 & .264 & .173 & .305 & .623 & .409 \\
Kikuyu     & .098 & .068 & .080 & .268 & .185 & .219 & .588 & .407 & .481 \\
Luo        & .131 & .091 & .107 & .285 & .198 & .233 & .637 & .441 & .521 \\
Xichangana & .112 & .108 & .110 & .280 & .270 & .275 & .460 & .444 & .451 \\
\midrule
\textbf{Macro Avg} & \textbf{.095} & \textbf{.091} & \textbf{.090} & \textbf{.232} & \textbf{.223} & \textbf{.219} & \textbf{.490} & \textbf{.475} & \textbf{.464} \\
\bottomrule
\end{tabular}
\caption{LLM judge vs.\ human ground truth (averaged across 3 runs). P = Precision, R = Recall, F1 = F1 score.}
\label{tab:model-vs-gt}
\end{table}

\subsubsection{Analysis}

The large gap between strict matching (F1 = 0.09) and type matching (F1 = 0.46) reveals that the LLM judge identifies relevant cultural \textit{categories} at moderate accuracy but differs substantially from humans in span boundary selection. Partial matching (F1 = 0.22) confirms that span overlap is limited. We identify three primary sources of disagreement:

\begin{itemize}
    \item \textbf{Span granularity mismatch.} Human annotators typically mark short, specific phrases (e.g., ``fried dough'', ``morning prayers''), while the LLM judge tends to extract longer, sentence-level spans that encompass the cultural marker along with its surrounding context. This systematically reduces strict and partial match scores without necessarily reflecting a disagreement about cultural content.
    \item \textbf{Span mismatch.} Although the total number of model predictions (1,134--1,701 per language) is sometimes comparable to the human ground truth (781--2,453), the model's spans frequently do not align with human annotations. For AmaXhosa, the model produces roughly twice as many spans as the ground truth (1,630 vs.\ 781), while for Luo and Kikuyu the model produces fewer spans than the human ground truth (1,636 vs.\ 2,453 and 1,479 vs.\ 2,068 respectively). In all cases, the low precision and recall scores indicate that model and human spans identify different portions of text as culturally relevant, even when they agree on the category.
    \item \textbf{Category confusion.} Even when spans partially overlap, the model and humans sometimes assign different cultural categories. The type-level F1 (0.41--0.52) indicates that approximately half of the model's category assignments align with human judgments. Category confusion is most pronounced for broad categories such as \textit{place\_physical\_environment} and \textit{occupation\_daily\_routine}, which the model tends to over-predict at the expense of more specific categories.

    \item \textbf{Ceiling effects.} Human inter-annotator agreement on span boundaries (F1 = 0.42) provides context for interpreting model performance. The model's type-level F1 (0.46) approaches human category agreement (0.67), suggesting that while boundary alignment remains challenging, the model's category-level understanding is within reach of human performance when boundary constraints are relaxed.
    
\end{itemize}

In general the performance varies across languages, Luo achieves the highest strict F1 (0.107) and type precision (0.637), while AmaXhosa shows the lowest strict F1 (0.072) but highest type recall (0.623). These differences may reflect variation in annotation density, category distribution, or the degree to which cultural markers in each language are expressed through discrete, identifiable phrases versus diffuse narrative context.

\end{document}